\documentclass{article}

\usepackage{iclr2027_conference,times}

\usepackage[english]{babel}
\usepackage{bold-extra}
\usepackage{bbold}
\usepackage{bm}
\usepackage{amsmath,amsthm,amssymb}
\usepackage{adjustbox}
\usepackage[dvipsnames]{xcolor}
\definecolor{tyellow1}{HTML}{FCE94F}
\definecolor{tyellow2}{HTML}{EDD400}
\definecolor{tyellow3}{HTML}{C4A000}
\definecolor{torange1}{HTML}{FCAF3E}
\definecolor{torange2}{HTML}{F57900}
\definecolor{torange3}{HTML}{C35C00}
\definecolor{tbrown1}{HTML}{E9B96E}
\definecolor{tbrown2}{HTML}{C17D11}
\definecolor{tbrown3}{HTML}{8F5902}
\definecolor{tgreen1}{HTML}{8AE234}
\definecolor{tgreen2}{HTML}{73D216}
\definecolor{tgreen3}{HTML}{4E9A06}
\definecolor{tblue1}{HTML}{729FCF}
\definecolor{tblue2}{HTML}{3465A4}
\definecolor{tblue3}{HTML}{204A87}
\definecolor{tpurple1}{HTML}{AD7FA8}
\definecolor{tpurple2}{HTML}{75507B}
\definecolor{tpurple3}{HTML}{5C3566}
\definecolor{tred1}{HTML}{EF2929}
\definecolor{tred2}{HTML}{CC0000}
\definecolor{tred3}{HTML}{A40000}
\definecolor{tlgray1}{HTML}{EEEEEC}
\definecolor{tlgray2}{HTML}{D3D7CF}
\definecolor{tlgray3}{HTML}{BABDB6}
\definecolor{tdgray1}{HTML}{888A85}
\definecolor{tdgray2}{HTML}{555753}
\definecolor{tdgray3}{HTML}{2E3436}

\definecolor{cgreen}{HTML}{008000}
\definecolor{rblue}{HTML}{0067A5}

\usepackage[%pdftex%
,colorlinks=true       % false: boxed links; true: colored links
,linkcolor=rblue       % color of internal links
,citecolor=cgreen      % color of links to bibliography
,filecolor=black       % color of file links
,urlcolor=rblue        % color of external links
,linktoc=page          % only page is linked
,plainpages=false
,pdftitle={On Formal Counterfactual Explanations}
,pdfauthor={Alexey Ignatiev, Peter J. Stuckey, Joao Marques-Silva}]{hyperref}
\addto\extrasenglish{%

}
\usepackage[utf8]{inputenc} % allow utf-8 input
\usepackage[T1]{fontenc}    % use 8-bit T1 fonts
\usepackage[algo2e,noend,ruled,noline]{algorithm2e}
\usepackage{algorithm}
\usepackage{amsfonts,amsmath,latexsym,nicefrac,eufrak,xspace}
\usepackage{empheq}
\usepackage{extarrows}
\usepackage{caption}
\usepackage{subcaption}
\usepackage{adjustbox,soul}
\usepackage{fancybox}
\usepackage{multirow,siunitx}
\usepackage{comment}
\usepackage{booktabs}
\usepackage{stmaryrd}
\graphicspath{{plots/}}
\DeclareGraphicsExtensions{.pdf,.png}

\usepackage{tikz}
\usetikzlibrary{arrows,calc,decorations,fit,positioning,shapes,tikzmark}

\usepackage{listings}
\newtheorem{proposition}{Proposition}
\newtheorem{lemma}{Lemma}
\newtheorem{definition}{Definition}
\newtheorem{example}{Example}

\AtBeginDocument{\renewcommand{\hbar}{\hslash}}

\newcommand{\fml}[1]{{\mathcal{#1}}}

\newcommand{\sbf}[1]{\ensuremath\mathbf{#1}}
\newcommand{\mbb}[1]{\ensuremath\mathbb{#1}}

\newcommand{\hs}{\textrm{HS}}
\newcommand{\mins}{\textrm{mins}}
\newcommand{\mhs}{\textrm{MHS}}

\newcommand{\axps}{\ensuremath\mbb{A}}
\newcommand{\cxps}{\ensuremath\mbb{B}}

\newcommand{\Zeta}{Z}

\def\squareforqed{\hbox{\rlap{$\sqcap$}$\sqcup$}}
\def\qed{\ifmmode\squareforqed\else{\unskip\nobreak\hfil
\penalty50\hskip1em\null\nobreak\hfil\squareforqed
\parfillskip=0pt\finalhyphendemerits=0\endgraf}\fi}

\colorlet{revcolor}{black}
\newcommand{\rev}[1]{{\color{revcolor}#1}}
\newenvironment{revblock}{\color{revcolor}}{}

\title{On Formal Counterfactual Explanations}

\author{%
  Alexey Ignatiev \\
  Monash University, Australia \\
  \texttt{alexey.ignatiev@monash.edu} \\
  \And
  Peter J.\ Stuckey \\
  Monash University, Australia \\
  \texttt{peter.stuckey@monash.edu} \\
  \And
  Joao Marques-Silva \\
  ICREA \& IIIA-CSIC, Spain \\
  \texttt{jpms@icrea.cat} \\
}

\iclrfinalcopy % Uncomment for camera-ready version, but NOT for submission.
\begin{document}

% \pagestyle{plain}
% \date{} % no date

\maketitle

\begin{abstract}
  Counterfactual explanations are often seen as counterfactual
  conditionals under \emph{closest-world} semantics, requiring one to
  reason over all minimally distant counterexamples.
  In contrast, formal XAI typically relies on \emph{feature-based}
  abductive explanations (AXps) and contrastive explanations (CXps),
  which admit efficient computation and a minimal hitting-set duality.
  This paper bridges these perspectives by (i)~characterizing
  closest-world counterfactual explanations explicitly as a
  disjunctive normal form over all minimally distant counterexample
  instances, and (ii)~introducing \emph{conjunctive counterfactual
  explanations} (CFXps), a subclass of counterfactual antecedents
  restricted to conjunctions of feature literals that guarantee
  prediction change.
  The paper establishes structural results connecting CXps, AXps and
  CFXps: every CXp induces an AXp for its witness instance, and every
  CXp together with a witness AXp induces a CFXp whose support is
  precisely determined by that AXp.
  These theoretical results yield a modular approach for computing
  CFXps using existing AXp/CXp machinery.
  Experiments on a representative set of benchmarks using tree
  ensembles and neural networks show that CFXp extraction is
  practically efficient.
  Compared with popular point-based counterfactual explainers,
  CFXps generally require no more feature changes and can guarantee the
  counterfactual prediction beyond a single instance.
\end{abstract}

\section{Introduction} \label{sec:intro}

Counterfactual reasoning plays a central role in human explanation and
decision-making~\citep{miller-aij19}.
When faced with an outcome, people naturally ask: \emph{``what would
need to change for the decision to be different?''}.
Recent work~\citep{lorini-logcom23} formalizes this intuition using a
modal-logic view of counterfactual conditionals, grounded in
closest-world semantics of
Stalnaker-Lewis~\citep{stalnaker-theory68,lewis-counterfactuals73}.
In parallel, formal explainable AI (FXAI) has largely focused on two
complementary types of feature-based explanations: abductive
explanations (AXps), which justify a prediction, and contrastive
explanations (CXps), which identify changes that can alter
it~\citep{inams-aiia20}.
AXps and CXps enjoy a minimal hitting set duality~\citep{inams-aiia20},
and admit exact computation for a wide range of
classifiers~\citep{msi-aaai22}.

Stalnaker-Lewis' counterfactuals~\citep{lorini-logcom23} reason
universally over closest counterexamples, whereas CXps require only
one~\citep{miller-aij19,inams-aiia20}.
The former allow arbitrary propositional antecedents, while AXps
and CXps are restricted to conjunctions of feature-literals.
This mismatch creates computational and representational challenges:
closest-world counterfactuals can be exponentially (or even
infinitely) large and need not correspond to a single feature subset.

We bridge these perspectives by distinguishing two kinds of
counterfactual explanations.
First, we capture Stalnaker-Lewis' closest-world semantics by a formal
closest-world counterfactual explanation: a formula in disjunctive
normal form (DNF) over all minimally distant counterexample instances.
Second, and more importantly, we introduce an efficiently computable
and actionable subclass of formal feature-based counterfactuals, which
we call \emph{conjunctive counterfactual explanations}.
Like AXps and CXps, they are feature subsets corresponding to
conjunctions of feature-literals, but unlike CXps they guarantee,
rather than merely enable, a prediction change.
We study their properties, relation with AXps and CXps, and
computation.

Our main contributions are as follows.
First, we formalize Stalnaker-Lewis closest-world counterfactuals in
the FXAI setting, building on recent modal-logic
work~\citep{lorini-logcom23}, and discuss their computation.
Second, we introduce \emph{conjunctive} counterfactual explanations
(CFXps), a syntactic and computationally tractable form of
counterfactual antecedent that blends the existential semantics of
CXps with the universal semantics of AXps.
Their definition requires neither a distance function nor finite
feature domains.
Third, we prove two structural results relating CXps, AXps, and CFXps:
(1)~every CXp induces an AXp for its witness instance, and (2)~every
CXp induces a CFXp whose support set is precisely characterized by the
witness' AXp.
Finally, these results yield an efficient and modular algorithm for
computing CFXps using the existing FXAI machinery.
We evaluate CFXp extraction on XGBoost tree ensembles and ReLU neural
networks using independent MaxSAT and MILP backends, and compare it
with popular point-based counterfactual explainers.
Together, these contributions provide a formal counterfactual
reasoning framework and clarify the relations between abductive,
contrastive, and conjunctive counterfactual explanations.

\section{Preliminaries} \label{sec:prelim}

\subsection{Classification Problems}

Let $\varphi=\{1,\ldots,m\}$, $m\in\mbb{N}$, be a set of
\emph{features}.
Subsets of $\varphi$ will also be denoted by lower-case Greek letters,
such as $\alpha$ and $\beta$.
A feature $j\in\varphi$ is characterized by a set of possible values,
called its \emph{domain} $\mbb{D}_j$.
Domain $\mbb{D}_j$ can be \emph{totally ordered}, in which case
feature $j$ is \emph{ordinal} or \emph{numerical}; otherwise, feature
$j$ is said to be \emph{categorical}.
The \emph{complete feature space} is defined by
$\mathbb{F}\triangleq\prod_{j=1}^{m}\mbb{D}_j$.
A \emph{data point} in $\mbb{F}$ is denoted $\sbf{v} = (v_1, \ldots,
v_m)$, where $v_j \in \mbb{D}_j$, $j\in\varphi$.
Also, we use $\sbf{x}=(x_1,\ldots,x_m)$ to denote a vector variable in
$\mbb{F}$, where coordinate $x_j$ stores
the~value~of~feature~$j$~from~$\mbb{D}_j$.

We will also need the notion of \emph{compatible subspaces}.
Given a subset of features $\zeta\subseteq\varphi$, define an
\emph{equivalence} relation $\sim_\zeta$ on $\mbb{F}$:
% for any two
% data points $\sbf{x},\sbf{y}\in\mbb{F}$, we say that points $\sbf{x}$
% and $\sbf{y}$ are equivalent on the set of features $\zeta$ denoted as
$\sbf{x}\sim_\zeta\sbf{y}$ if and only if $x_j=y_j$, $\forall
j\in\zeta$.
%
% In other words, such points $\sbf{x}$ and $\sbf{y}$ are equivalent on
% $\zeta$ iff they share the values of all features $j\in\zeta$.
%
Given a point $\sbf{v}\in\mbb{F}$, relation
$\sim_\zeta$ defines the equivalence class $[\sbf{v}]_\zeta \triangleq
\{\sbf{x} \in \mbb{F} \mid \sbf{x} \sim_\zeta \sbf{v}\}$.
Moreover, $\sim_\zeta$ splits $\mbb{F}$ into
$\lvert\prod_{j\in\zeta}\mbb{D}_j\rvert$ equivalence classes, for
any $\zeta\in 2^\varphi$.
We refer to $[\sbf{v}]_\zeta$ as the subspace of $\mbb{F}$ compatible
with $\sbf{v}$ on $\zeta$.

\begin{lemma} \label{th:lemma}
  For any $\sbf{v} \in \mbb{F}$ and feature subsets
  $\zeta_1 \subseteq \varphi$ and $\zeta_2 \subseteq \varphi$,
  % such that $\zeta_1 \cap \zeta_2 \neq \emptyset$
  it holds that
  \begin{equation} \label{eq:cap}
    [\sbf{v}]_{\zeta_1} \cap [\sbf{v}]_{\zeta_2} \equiv
    [\sbf{v}]_{\zeta_1 \cup \zeta_2}
  \end{equation}
\end{lemma}

\begin{proof}[Proof (Sketch)]
  %
  % By definition, given a point $\sbf{v} \in \mbb{F}$, relations
  % $\sim_{\zeta_1}$ and $\sim_{\zeta_2}$ determine two equivalence
  % classes
  % %
  % $[\sbf{v}]_{\zeta_1} = \{\sbf{x} \in \mbb{F} \mid \sbf{x}
  % \sim_{\zeta_1} \sbf{v}\}$
  % %
  % and
  % %
  % $[\sbf{v}]_{\zeta_2} = \{\sbf{x} \in \mbb{F} \mid \sbf{x}
  % \sim_{\zeta_2} \sbf{v}\}$.
  % %
  % Thus, $[\sbf{v}]_{\zeta_1} \cap [\sbf{v}]_{\zeta_2} = \{\sbf{x} \in
  % \mbb{F} \mid \sbf{x} \sim_{\zeta_1} \sbf{v} \,\land\, \sbf{x}
  % \sim_{\zeta_2} \sbf{v}\}$, i.e.\ their intersection comprises points
  % in the feature space that are compatible with $\sbf{v}$ \emph{both}
  % on the subset of features $\zeta_1$ \emph{and} the subset of features
  % $\zeta_2$.
  % %
  % Recall that $\sbf{x} \sim_\alpha \sbf{v}$ iff $x_j=v_j$, $\forall j
  % \in \alpha$.
  % %
  % So $[\sbf{v}]_{\zeta_1} \cap [\sbf{v}]_{\zeta_2} = \{ \sbf{x} \in
  % \mbb{F} \mid x_j = v_j, \forall j \in \zeta_1 \,\land\, x_i = v_i,
  % \forall i \in \zeta_2\} = \{ \sbf{x} \in \mbb{F} \mid x_j = v_j,
  % \forall j \in \zeta_1 \cup \zeta_2 \} = \{ \sbf{x} \in \mbb{F}
  % \mid \sbf{x} \sim_{\zeta_1 \cup \zeta_2} \sbf{v}\} =
  % [\sbf{v}]_{\zeta_1 \cup \zeta_2}$.

  By definition, a point $\sbf{x}\in\mbb{F}$ belongs to
  $[\sbf{v}]_{\zeta_1} \cap [\sbf{v}]_{\zeta_2}$ iff it agrees with
  $\sbf{v}$ on \emph{all} features in both subsets $\zeta_1$ and
  $\zeta_2$, i.e. on all features in $\zeta_1\cup\zeta_2$.
  Hence $[\sbf{v}]_{\zeta_1} \cap [\sbf{v}]_{\zeta_2} =
  [\sbf{v}]_{\zeta_1\cup\zeta_2}$.
  %
  % \qed
  %
\end{proof}

Identity \eqref{eq:cap} is used later.
Notably, $[\sbf{v}]_\zeta$ is simple to represent logically using the
conjunction of literals $\bigwedge_{i\in\zeta}\left(x_i=v_i\right)$.
An assignment of values to variables $x_i\in\mbb{D}_i$, $i\in\varphi$,
trivially satisfies this conjunction iff the data point
$\sbf{x}=(x_1,\ldots,x_m)$ belongs to the compatible subspace, i.e.
$\sbf{x}\in[\sbf{v}]_\zeta$.

We consider a \emph{total}, \emph{surjective} classification function
$\kappa: \mathbb{F} \rightarrow \mbb{K}$,
i.e.\ each point $\sbf{x}\in\mbb{F}$ has an associated class
$c\in\mbb{K}$ and, vice versa, each class \emph{``contains''} at least
one point mapped to it by $\kappa$.\footnote{%
  The ideas also extend to regression problems.
}
A classification problem is characterized by the set of features
$\varphi$, feature space $\mbb{F}$, set of classes $\mbb{K}$, and the
classification function $\kappa$.
Formally, a machine learning (ML) classifier is the tuple
$M=(\varphi,\mbb{F},\mbb{K},\kappa)$.
Similar to truth tables in Boolean algebra, classification function
$\kappa$ can in principle be explicitly represented by listing all
data points $\sbf{v}\in\mbb{F}$ and the corresponding classes
$\kappa(\sbf{v})=c\in\mbb{K}$, although depending on the size
$\lvert\mbb{F}\rvert$ this may be \emph{infeasible} or highly
inefficient in practice.
A row in such a table represented by a pair $\sbf{i}=(\sbf{v},
c)\in\mbb{F}\times\mbb{K}$ such that $\kappa(\sbf{v})=c$ is referred
to as an \emph{instance}.
Finally, given a classifier $M$ and a concrete instance $\sbf{i}$, an
explanation problem is a tuple $\fml{E}=(M,\sbf{i})$.
Classifiers include decision trees~\citep{rivest-ipl76},
lists~\citep{rivest-ml87} and
sets~\citep{clark-ewsl91,lakkaraju-kdd16}, random
forests~\citep{friedman-tas01}, boosted trees~\citep{guestrin-kdd16a},
and neural nets~\citep{hinton-icml10,hcseyb-neurips16}.

% \begin{example} \label{ex:models}
%   %
%   Given the well-known \emph{Iris} dataset aiming to classify the
%   three species of Iris (``setosa'', ``versicolor'', and
%   ``virginica''), example decision set and boosted treed models are
%   shown in \autoref{fig:models}.
%   %
%   Based on the four features: (1)~sepal length, (2)~sepal width,
%   (3)~petal length, and (4)~petal width, the models classify an input
%   instance as one of the three species.
%   %
%   Importantly, both models compute exactly the same classification
%   function and so behave \emph{identically} on all the instances in
%   the feature space.
%   %
%   \qed
%   %
% \end{example}

\begin{example} \label{ex:toy1}
  Consider classification function $\kappa: \{0,1\}^3 \rightarrow
  \{0,1\}$:
  \begin{equation*}
    \kappa(x_1,x_2,x_3) \;=\;
    \begin{cases}
      1 & \text{ if } (x_1 \land x_2) \lor x_3 \\
      0 & \text{ otherwise }
    \end{cases}
  \end{equation*}
  The function can be fully specified using the truth table shown in
  \autoref{tab:toy1func}.
  In the following, we will consider the point
  $\sbf{v}=\sbf{v}_5=(1,0,0)$ classified by $\kappa$ as $0$.
\end{example}

% \begin{example} \label{ex:toy2}
%   %
%   Consider function $\kappa_2: \{0,1,2\} \times \{0,1\}
%   \times \{0,1\} \rightarrow \{0,1,2\}$:
%   %
%   \begin{equation*}
%     %
%     \kappa_2(x_1,x_2,x_3) \;=\;
%     %
%     \begin{cases}
%       %
%       0 & \text{ if } x_1 = 0  \text{, or } x_1 = 1 \text{ and } x_2 = 0 \text{, or } x_1=1 \text{ and } x_3 = 0 \\
%       %
%       1 & \text{ if } x_1 = 1 \text{ and } x_2 = 1 \\
%       %
%       2 & \text{ if } x_1 = 2 \text{ and } x_3 = 1
%       %
%     \end{cases}
%     %
%   \end{equation*}
%   %
%   % This setup can represent a loan decision system such that feature
%   % $1$ represents one's \emph{income}, which can be low, medium, and
%   % high; feature $2$ represents one's \emph{credit score} with the
%   % values low and high; and feature $3$ represents one's
%   % \emph{savings}, which can also be low or high.
%   % %
%   % The class values $0$, $1$, and $2$ can act as the predictions
%   % \emph{rejected}, \emph{manual review}, and \emph{accepted}.
%   % %
%   % \autoref{fig:toy2dt} depicts a decision tree (DT) model that
%   % computes classification function $\kappa_2$, using the above
%   % natural-language feature names and values.
%   %
%   This classification function can be fully specified using the truth
%   table shown in \autoref{tab:toy2func}.
%   %
%   In the following examples, we will consider a point $\sbf{u} \equiv
%   \sbf{u}_1=(0, 0, 0)$, which is classified by $\kappa_2$ as $0$.
%   %
%   \qed
%   %
% \end{example}

\begin{table}[t!]
  \caption{Truth table for the classification function from
  \autoref{ex:toy1}.
  %
  % The instance of interest is $\sbf{v}_5$.
  %
  }
  \label{tab:toy1func}
  \centering
  \setlength{\tabcolsep}{6pt}
  \begin{tabular}{ccccccccccccccc}
    \toprule
    && $\sbf{v}_1$ & $\sbf{v}_2$ & $\sbf{v}_3$ & $\sbf{v}_4$ &
    $\sbf{v}_5$ & $\sbf{v}_6$ & $\sbf{v}_7$ & $\sbf{v}_8$ \\
    \midrule
    $x_1$    && $0$ & $0$ & $0$ & $0$ & $1$ & $1$ & $1$ & $1$ \\
    $x_2$    && $0$ & $0$ & $1$ & $1$ & $0$ & $0$ & $1$ & $1$ \\
    $x_3$    && $0$ & $1$ & $0$ & $1$ & $0$ & $1$ & $0$ & $1$ \\
    \midrule
    $\kappa(\sbf{x})$ && $0$ & $1$ & $0$ & $1$ & $0$ & $1$ & $1$ & $1$
    \\
    \bottomrule
  \end{tabular}
\end{table}

% \begin{table}[t]
%   %
%   \caption{Truth table for the classification function from
%   \autoref{ex:toy2}.
%   %
%   % The instance of interest is $\sbf{u}_1$.
%   %
%   }
%   %
%   \label{tab:toy2func}
%   %
%   \centering
%   %
%   \setlength{\tabcolsep}{6pt}
%   %
%   \begin{tabular}{ccccccccccccccccccc}
%     %
%     \toprule
%     %
%     && $\sbf{u}_1$ & $\sbf{u}_2$ & $\sbf{u}_3$ & $\sbf{u}_4$ &
%     $\sbf{u}_5$ & $\sbf{u}_6$ & $\sbf{u}_7$ & $\sbf{u}_8$ &
%     $\sbf{u}_9$ & $\sbf{u}_{10}$ & $\sbf{u}_{11}$ & $\sbf{u}_{12}$ \\
%     %
%     %
%     \midrule
%     %
%     $x_1$    && $0$ & $0$ & $0$ & $0$ & $1$ & $1$ & $1$ & $1$ & $2$ & $2$ & $2$ & $2$ \\
%     %
%     $x_2$    && $0$ & $0$ & $1$ & $1$ & $0$ & $0$ & $1$ & $1$ & $0$ & $0$ & $1$ & $1$ \\
%     %
%     $x_3$    && $0$ & $1$ & $0$ & $1$ & $0$ & $1$ & $0$ & $1$ & $0$ & $1$ & $0$ & $1$ \\
%     %
%     \midrule
%     %
%     $\kappa_2(\sbf{x})$ && $0$ & $0$ & $0$ & $0$ & $0$ & $0$ & $1$ & $1$ & $0$ & $2$ & $0$ & $2$ \\
%     %
%     \bottomrule
%     %
%   \end{tabular}
%   %
% \end{table}

% \begin{figure*}[t]
%   %
%   \begin{adjustbox}{center}
%     %
%     \input{texfigs/toy2.tex}
%     %
%   \end{adjustbox}
%   %
%   \caption{Decision tree computing the classification function of
%   \autoref{ex:toy2}.}
%   %
%   \label{fig:toy2dt}
%   %
% \end{figure*}

\subsection{Interpretability \& Formal Explanations} \label{sec:exp}
Interpretability is generally accepted to be a subjective concept,
without a formal definition~\citep{lipton-cacm18}.
In this paper, we view it in terms of the succinctness of the
information used by a classifier to justify a prediction.
Building on earlier work, we equate explanations with
the so-called \emph{abductive explanations}
(AXps)~\citep{darwiche-ijcai18,inms-aaai19,inms-nips19,darwiche-ecai20,iims-corr20,inams-aiia20,barcelo-nips20,msgcin-nips20,msgcin-icml21,hiims-kr21,msi-aaai22,hiicams-aaai22},
i.e.\ subsets of features that are sufficient for the prediction.
Formally, given an explanation problem $\fml{E}=(M,\sbf{i})$ with
classifier $M=(\varphi,\mbb{F},\mbb{K},\kappa)$ and instance
$\sbf{i}=(\sbf{v},c)$, a \emph{weak AXp} is a subset of
features $\alpha\subseteq\varphi$ such that,
\begin{equation} \label{eq:axp}
  \forall \sbf{x}\in[\sbf{v}]_\alpha. \quad \kappa(\sbf{x})=c \\
\end{equation}
Weak AXps capture the \emph{selectis
paribus} principle (``selected things being
equal'')~\citep{lorini-logcom23}: all points in the subspace
of $\mbb{F}$ compatible with point $\sbf{v}$ on $\alpha$ are
classified as class $c$.
Using the logical formulation of compatible subspaces, \eqref{eq:axp}
can be written as:
%
% Using the logical formulation of the compatible subspace
% $[\sbf{v}]_\alpha$, definition \eqref{eq:axp} can be represented
% logically using the following predicate, which is directly amenable to
% deciding by a formal reasoning engine:
%
\begin{equation} \label{eq:axpred}
  \forall \sbf{x}\in\mbb{F}. \quad \left[\bigwedge_{i\in\alpha}
  \left(x_i=v_i\right) \right] \rightarrow
  \left(\kappa(\sbf{x})=c\right) \\
\end{equation}

Following Occam's razor, a feature subset
$\alpha\subseteq\varphi$ is an abductive explanation (AXp, or strong
AXp) if (i)~$\alpha$ is a weak AXp and (ii)~$\alpha$ is
subset-minimal, i.e.\ no \emph{proper} subset of $\alpha$ is a weak
AXp.
Some works also refer to AXps as \emph{prime implicant} explanations
(or PI-explanations)~\citep{darwiche-ijcai18} or \emph{sufficient
reasons}~\citep{darwiche-ecai20}.

Similarly, a \emph{contrastive explanation}
(CXp)~\citep{miller-aij19,inams-aiia20} is a feature subset that enables
the prediction change when its features take \emph{arbitrary} values
from their domains.
Formally, a \emph{weak CXp} is a feature subset
$\beta\subseteq\varphi$ such that,
\begin{equation} \label{eq:cxp}
  \exists \sbf{x}\in[\sbf{v}]_{\varphi\setminus\beta}.\quad
  \kappa(\sbf{x})\neq c
\end{equation}
Unlike AXps, weak CXps capture the \emph{ceteris paribus}
principle (``all other things being equal''): there is at least
one point in the subspace of $\mbb{F}$ compatible with $\sbf{v}$ on
$\varphi\setminus\beta$ that is classified differently from
class $c$.
Using the logical formulation of compatible subspaces,
\eqref{eq:cxp} can be written as:
\begin{equation} \label{eq:cxpred}
  \exists \sbf{x}\in\mbb{F}. \quad
  \left[\bigwedge_{i\in\varphi\setminus\beta} \left(x_i=v_i\right)
  \right] \land \left(\kappa(\sbf{x})\neq c\right) \\
\end{equation}
A feature subset $\beta\subseteq\varphi$ is a contrastive explanation
(CXp, or strong CXp) if it is a subset-minimal weak CXp.
CXps are also referred to as \emph{necessary
reasons}~\citep{darwiche-aaai22}.
% (It is possible and simple to adapt the definition to target a
% specific class $c'\not=c$.)

% Motivated by the relationship between diagnoses and explanations in
% the context of model-based diagnosis~\citep{reiter-aij87}, recent work
% demonstrated that AXps and CXps enjoy a minimal hitting set duality
% relationship~\citep{inams-aiia20}, namely each AXp is a minimal hitting
% set (MHS) of the set of CXps and vice-versa.
%
Recent work showed that AXps and CXps satisfy a minimal hitting set
(MHS) duality~\citep{inams-aiia20}: each AXp is a minimal hitting set
of the set of CXps, and vice versa.
Given an explanation problem $\fml{E}=(M,\sbf{i})$ characterized by a
classifier $M$ and a particular instance $\sbf{i}=(\sbf{v},c)$, let
$\axps_\fml{E}$ denote the complete set of AXps and $\cxps_\fml{E}$
denote the complete set of CXps.
Then $\axps_\fml{E}=\mhs(\cxps_\fml{E})$ and
$\cxps_\fml{E}=\mhs(\axps_\fml{E})$, where $\mhs(S)$ returns all
minimal hitting sets of $S$, i.e.\ minimal sets that share an element
with each subset in $S$.
Formally, $\hs(S)=\{t\subseteq(\cup S)\mid \forall s\in S,\ t\cap
s\neq\emptyset\}$, $\mins(S)=\{s\in S\mid \forall t\subsetneq s,\
t\notin S\}$ returns the subset-minimal elements of a set of sets, and
$\mhs(S)=\mins(\hs(S))$.
This MHS duality has enabled efficient algorithms for computing and
enumerating AXps and CXps~\citep{inams-aiia20,msi-aaai22}.

% \begin{example} \label{ex:xps}
%   %
%   Consider any model from \autoref{ex:models} and a data point
%   $\sbf{v}=(5.1,3.5,1.4,0.2)$. (Recall that \emph{sepal.length},
%   \emph{sepal.width}, \emph{petal.length}, and \emph{petal.width} are
%   features 1, 2, 3, and 4, respectively.)
%   %
%   Observe that either of the models outputs class ``setosa''.
%   %
%   Also note that it suffices to specify $\text{\em
%   petal.length}=\text{\em 1.4}$ to guarantee this prediction.
%   %
%   Therefore, a valid abductive explanation for this instance includes
%   a single feature \emph{petal.length}, i.e.\ $\alpha=\{3\}$, because
%   there is no point in $[\sbf{v}]_{\{3\}}$ that could be predicted
%   differently than ``setosa''.
%   %
%   Furthermore, it is the only AXp for the given instance.
%   %
%   Similarly, a valid contrastive explanation is $\beta=\{3\}$, because
%   there exists a point in $[\sbf{v}]_{\{1,2,4\}}$ predicted
%   differently, e.g.\ $(5.1,3.5,\ast 2.6\ast,0.2)$.
%   %
%   Finally, observe that $\axps_\fml{E}=\{\{3\}\}$ and
%   $\cxps_\fml{E}=\{\{3\}\}$ illustrating the minimal hitting set
%   duality.
%   %
%   \qed
%   %
% \end{example}

\begin{example} \label{ex:toy1xp}
  Consider the explanation problem $\fml{E}=(M,\sbf{i})$, with
  classifier $M$ characterized by the features $\varphi=\{1,2,3\}$,
  feature space $\mbb{F}=\{0,1\}^3$, classes $\mbb{K}=\{0,1\}$, as
  well as classification function $\kappa(x_1, x_2, x_3)$ and the
  instance $\sbf{i}=(\sbf{v}_5,0)$, $\sbf{v}_5=(1, 0, 0)$ from
  \autoref{ex:toy1}.
  By inspection of \autoref{tab:toy1func}, the only AXp is
  $\alpha=\{2,3\}$, since fixing $x_2=0$ and $x_3=0$ guarantees class
  $0$.
  The two CXps are $\beta_1=\{2\}$ and $\beta_2=\{3\}$, witnessed by
  points $\sbf{v}_7=(1,1,0)$ and $\sbf{v}_6=(1,0,1)$, respectively.
  Hence $\axps_{\fml{E}}=\{\{2,3\}\}$ and
  $\cxps_{\fml{E}}=\{\{2\},\{3\}\}$, illustrating the hitting-set
  duality.
\end{example}

Related explanation problems have been studied for compiled
representations and Boolean classifier
families~\citep{marquis-kr20,marquis-corr21,darwiche-jair21}.
Other extensions incorporate background knowledge in the form of
domain constraints~\citep{rubin-aaai22,cooper-ecai23}, including
knowledge extracted from data~\citep{yu-aaai23}, or generalize AXps by
associating them with Craig
interpolants~\citep{craig-jsl57,mcmillan-cav03,sharygina-cav25}.
Formal guarantees further cover robust NLP
explanations~\citep{kwiatkowska-ijcai21}, approximate formal neural
explanations~\citep{bassan-tacas23}, and counterfactual anomaly
repair~\citep{ji-neurips24}.
Complementary axiomatic and complexity-theoretic accounts study
general properties of counterfactual and formal
explanations~\citep{amgoud-aamas26,bassan-iclr26}.

\rev{Many existing \emph{counterfactual point} explainers build on
  heuristic search or distance
  optimization~\citep{artelt-corr19,karimi-acmcs23,guidotti-dmkd24,verma-acmcs24}.
  Some methods~\citep{guidotti-corr18} extract regions from a local
  surrogate while others~\cite{upadhyay-neurips21} rely on
  perturbations, thus offering no guarantees for the classifier.
  In contrast, the CFXps proposed here formally guarantee a
  counterfactual prediction of the classifier over a whole region.
}

% \begin{example} \label{ex:toy2xp}
%   %
%   Consider the explanation problem $\fml{E}_2=(M_2,\sbf{i}_2)$, s.t.\
%   classifier $M_2$ is characterized by the set of features
%   $\varphi=\{1,2,3\}$, feature space $\mbb{F}=\{0,1,2\} \times \{0,1\}
%   \times \{0,1\}$, set of classes $\mbb{K}=\{0,1,2\}$, classification
%   function $\kappa_2(x_1, x_2, x_3)$ from \autoref{ex:toy2} and the
%   instance $\sbf{i}_2=(\sbf{u}_1,0)$, $\sbf{u}_1=(0, 0, 0)$.
%   %
%   The two AXps for this instance are: $\alpha_1=\{1\}$ and
%   $\alpha_2=\{2,3\}$.
%   %
%   Indeed, keeping $x_1=0$ or both $x_2=0$ and $x_3=0$ guarantees the
%   prediction stays $0$.
%   %
%   The two CXps are: $\beta_1=\{1,2\}$ and $\beta_2=\{1,3\}$.
%   %
%   The points in $\mbb{F}$ witnessing that $\beta_1$
%   and $\beta_2$ are CXps are $\sbf{u}_7=(1,1,0)$ predicted as
%   $1$ and $\sbf{u}_{10}=(2,0,1)$ predicted as $2$.
%   %
%   The minimal hitting set duality holds as $\axps_{\fml{E}_2}=\{\{1\},
%   \{2,3\}\}$ and $\cxps_{\fml{E}_2}=\{\{1,2\}, \{1,3\}\}$.
%   %
%   \qed
%   %
% \end{example}

\section{Formal Closest-World Counterfactuals} \label{sec:mlv}

This work builds on a recent modal-logic
account~\citep{lorini-logcom23}, which uses epistemic \emph{modal
logic} to define abductive and contrastive explanations via
\emph{necessity} ($\Box$) and \emph{possibility} ($\Diamond$).
It also relates counterfactual explanations for ML
models~\citep{miller-aij19,molnar-bk20} to \emph{counterfactual
conditionals} under closest-world semantics of
Stalnaker-Lewis~\citep{stalnaker-theory68,lewis-counterfactuals73}.

Informally, a counterfactual conditional $\tau\Rightarrow\rho$ holds
\emph{``if all closest worlds to the actual world in which the
antecedent $\tau$ is true satisfy the consequent $\rho$''}.
For an explanation problem $\fml{E}=(M,\sbf{i})$, the \emph{worlds}
represent data points $\sbf{x}\in\mbb{F}$ closest to the point of
interest $\sbf{v}\in\mbb{F}$, which represents the actual world.
Following~\citep[Proposition~3]{lorini-logcom23}, closest worlds can be
constructed using the Hamming distance between $\sbf{x}$ and
$\sbf{v}$.
% relative to a subset of features
% $\gamma\subseteq\varphi$.
% \footnote{%
%   %
%   Here, conditioning on the set of features $\gamma$
%   % is related to the concept of \emph{relativized
%   % conditionals}~\citep{girard-tark15} and
%   signifies which features are allowed to change.
%   %
%   As one is typically interested in taking into account all the
%   available features $\varphi$, conditioning can be omitted.
%   %
% }
%
For two points $\sbf{x}, \sbf{y} \in \mbb{F}$, their \emph{Hamming
distance} is the number of coordinates on which they disagree, i.e.\
\begin{equation*}
  d_H(\sbf{x}, \sbf{y})\triangleq \lvert \{i\in \varphi \mid x_i \neq
  y_i\}\rvert.
\end{equation*}

Here, the antecedent $\tau$ may be an arbitrary Boolean formula
serving as a counterfactual explanation for $\fml{E}=(M,\sbf{i})$,
$M=(\varphi,\mbb{F},\mbb{K},\kappa)$ and $\sbf{i}=(\sbf{v},c)$, while
the consequent $\rho$ of the conditional is meant to be the predicate
$\kappa_{\bar{c}}\triangleq\left(\kappa(\sbf{x})\neq c\right)$.
%
% In other words, a closest-world counterfactual explanation
% of~\citep{lorini-logcom23} is an arbitrary logical formula serving as
% the antecedent in the counterfactual conditional defined above, i.e.\
% if this antecedent formula is true on the set of such points $\sbf{x}$
% that are \emph{closest} to the point $\sbf{v}$ then another prediction
% $\kappa(\sbf{v}) = q \neq c$ is guaranteed.\footnote{The original work
% talks about maximizing the set of such closest points $\sbf{x}$.}
%
One natural choice for $\tau$ is a disjunctive normal form (DNF) whose
terms represent the closest points to $\sbf{v}$ where
$\kappa_{\bar{c}}$ holds.
Each term need only include literals on the features that
\emph{differentiate} the corresponding closest point from the point of
interest $\sbf{v}$.

\begin{example} \label{ex:toy1cf}
  For the point $\sbf{v}=\sbf{v}_5=(1,0,0)$, the Hamming-closest
  points are $\sbf{v}_1=(0,0,0)$, $\sbf{v}_6=(1,0,1)$, and
  $\sbf{v}_7=(1,1,0)$.
  Only $\sbf{v}_7$ and $\sbf{v}_6$ change the prediction, so the
  closest-world antecedent is $(x_2 \lor x_3)$, yielding $(x_2 \lor
  x_3) \Rightarrow (\kappa(\sbf{x})=1)$.
\end{example}

% \begin{example} \label{ex:toy2cf}
%   %
%   Consider the explanation problem $\fml{E}_2$ from \autoref{ex:toy2}.
%   %
%   The instance to be explained is $\sbf{u} \triangleq \sbf{u}_1 =
%   (0,0,0)$.
%   %
%   There are 4 points within Hamming distance $1$ from $\sbf{u}$,
%   including $\sbf{u}_2$, $\sbf{u}_3$, $\sbf{u}_5$, and $\sbf{u}_9$.
%   %
%   None of them is classified differently than $\sbf{u}$.
%   %
%   Extending the distance to $2$ enables us to reach a different
%   prediction for the points $\sbf{u}_7=(1,1,0)$ and
%   $\sbf{u}_{10}=(2,0,1)$, which are classified by $\kappa_2$ as $1$
%   and $2$, respectively.
%   %
%   Therefore, a counterfactual explanation of~\citep{lorini-logcom23} is
%   given as the antecedent of the counterfactual conditional
%   $\left[(x_1=1 \land x_2=1) \lor (x_1=2 \land x_3=1)\right]$
%   $\Rightarrow_{\{1,2,3\}} (\kappa_2(\sbf{x})\neq 0)$.
%   %
%   Here, the terms $x_1=1 \land x_2=1$ and $x_1=2 \land x_3=1$ in the
%   antecedent formula represent points $\sbf{u}_7$ and $\sbf{u}_{10}$,
%   respectively.
%   %
%   As a result, the counterfactual argument is to be understood as ``if
%   $x_1=1$ and $x_2=1$, or $x_1=2$ and $x_3=1$ while all other
%   coordinates \emph{being equal} to those of $\sbf{u}$, the prediction
%   will not be 0''.
%   %
%   \qed
%   %
% \end{example}

% Note how the closest-world counterfactual explanations capture the
% universal semantics of abductive explanations, i.e.\ \emph{as long
% as a given arbitrary formula holds true on the set of closest points
% to the point of interest, the class will be different}.
%
Closest-world counterfactual explanations thus capture universal
semantics akin to abductive explanations, i.e.\ \emph{if a formula
holds on all closest points, the class is guaranteed to differ}.
By contrast, contrastive explanations require only a witness for
prediction change.
%
% This is in clear contrast with contrastive explanations although
% both contrastive and counterfactual explanations relate with the
% change of the classifier's prediction.
% % One can envision various ways of devising such explanations
% formally. % In the following, we first discuss the general form of
% counterfactual explanations of~\citep{lorini-logcom23}, hereinafter
% referred to as \emph{closest-world (full) counterfactual
% explanations}, followed by a restricted but practically more
% appealing form of counterfactuals called \emph{conjunctive
% counterfactuals}.
%
% While~\citep{lorini-logcom23} does not propose algorithmic solutions
% for computing generic closest-world counterfactual explanations and
% the corresponding sets of closest points, we give such a solution
% below.
%
While~\citep{lorini-logcom23} gives no algorithm for these
explanations or their closest points, we give one below.
%
% Furthermore, next section will discuss a particular form of
% counterfactual explanation restricted to be a conjunction of feature
% literals $\bigwedge_{i \in \varphi \setminus \zeta}{(x_i=v_i)}$
% defined by a subset of features $\zeta \subseteq \varphi$, similar to
% the definition of an AXp and a CXp and an efficient way to compute it,
% targeting (cardinality- or) subset-minimality of such a subset.

For an explanation problem $\fml{E}=(M,\sbf{i})$ with
$M=(\varphi,\mbb{F},\mbb{K},\kappa)$ and
$\sbf{i}=(\sbf{v},c)$,~\citep{lorini-logcom23} considers the points
$\sbf{x} \in\mbb{F}$ with the \emph{smallest} Hamming distance from
$\sbf{v}$ such that $\kappa(\sbf{x}) \neq c$.
Let $d_H^{\min}(\sbf{v})=\min{\{d_H(\sbf{v},\sbf{x}) \mid \sbf{x} \in
\mbb{F}, \kappa(\sbf{x}) \neq c\}}$.
Then the set of Hamming-closest points to be used in the closest-world
counterfactual argument for $\sbf{v}$ can be expressed as
\begin{equation*}
  \mbb{E}_{\sbf{v}} \triangleq
  \{\sbf{x} \in \mbb{F} \mid d_H(\sbf{v}, \sbf{x})=d_H^{\min}(\sbf{v})
    \text{ and }
  \kappa(\sbf{x}) \neq c \}.
\end{equation*}
The set $\mbb{E}_{\sbf{v}}$ defines the DNF formula
\begin{equation} \label{eq:cmcf}
  \tau(\sbf{x}) \triangleq
  \bigvee_{\sbf{u} \in \mbb{E}_{\sbf{v}}}{\left[\bigwedge_{j \in
  \varphi}{(x_j = u_j)}\right]}
\end{equation}
serving as a counterfactual explanation for $\sbf{v}\in\mbb{F}$.
%
% The authors of~\citep{lorini-logcom23} argue that $\tau$ must comprise
% all the points in $\mbb{F}$ within a minimal distance from $\sbf{v}$
% that are predicted differently, hence we refer to it as a
% \emph{closest-world (full)} counterfactual explanation.
%
% Given that disjunctive representation \eqref{eq:cmcf}, one can also
% refer to it as a \emph{disjunctive counterfactual explanation}.
%
Since $\tau$ comprises all minimally distant points predicted
differently from $\sbf{v}$, we call \eqref{eq:cmcf} a
\emph{closest-world} or \emph{disjunctive} counterfactual explanation.

% For a point $\sbf{v}\in\mbb{F}$, computation of a closest-world
% counterfactual explanation requires one to enumerate the points in
% $\mbb{E}_{\sbf{v}}$.
%
Computing a closest-world counterfactual explanation $\mbb{E}_{\sbf{v}}$
requires enumerating all
\emph{cardinality-minimal} CXps $\beta\subseteq\varphi$ and all their
witnesses $\sbf{y} \in [\sbf{v}]_{\varphi \setminus \beta}$ such that
$\kappa(\sbf{y}) \neq c$.
%
% This exhaustive enumeration process is computationally expensive.
% %
% In fact, when $\mbb{F}$ is infinite with no succinct representation as
% a union of intervals, one cannot realistically complete it.
%
This exhaustive process is expensive and may be
infeasible when $\mbb{F}$ is infinite and lacks a succinct interval
representation.
%
% Additionally, given $\mbb{E}_{\sbf{v}}$ and assuming it is finite, one
% can also consider computing a minimal size DNF formula equivalent to
%
Additionally, if $\mbb{E}_{\sbf{v}}$ is finite, one may further seek a
minimum-size DNF equivalent to $\tau$ with an algorithm similar to the
Quine-McCluskey~\citep{quine-amm52,quine-amm55,mccluskey-bstj56}
procedure, thus solving the DNF minimization problem, which is
$\Sigma_2^P$-hard~\citep{umans-focs98,umans-tcad06}.

\begin{example} \label{ex:toy1dcfxp}
  For $\fml{E}=(M,(\sbf{v}_5,0))$, the smallest CXps are
  $\beta_1=\{2\}$ and $\beta_2=\{3\}$, witnessed by $\sbf{v}_7$ and
  $\sbf{v}_6$, respectively.
  Here $\tau(\sbf{x}) = (x_1 \land x_2 \land \neg{x_3}) \lor (x_1
  \land \neg{x_2} \land x_3)$. Dropping the literals shared with
  $\sbf{v}_5$ reduces it to $x_2 \lor
  x_3$~\citep{stalnaker-theory68,lewis-counterfactuals73,lorini-logcom23}.
\end{example}

% \begin{example} \label{ex:toy2dcfxp}
%   %
%   Recall the explanation $\fml{E}_2$.
%   %
%   As was established in~\autoref{ex:toy2xp}, there are two smallest
%   size CXps for the point $\sbf{u}=(0,0,0)$, namely $\beta_1=\{1,2\}$
%   and $\beta=\{1,3\}$ witnessed by the points $\sbf{u}_7=(1,1,0)$ and
%   $\sbf{u}_{10}=(2,0,1)$, respectively.
%   %
%   As such, the DNF formula representing the closest-world
%   counterfactual explanation is $\tau(\sbf{x}) = (x_1=1 \land x_2=1
%   \land x_3=0) \lor (x_1=2 \land x_2=0 \land x_3=1)$.
%   %
%   Under Stalnaker-Lewis semantics, we can opt to explicitly show only
%   the literals different from those in point $\sbf{u}=(0,0,0)$, which
%   results in the sought antecedent of the counterfactual conditional,
%   i.e.\ $\left[(x_1=1 \land x_2=1) \lor (x_1=2 \land x_3=1)\right]$.
%   %
%   \qed
%   %
% \end{example}

\section{Formal Conjunctive Counterfactuals} \label{sec:fcc}

The closest-world counterfactual explanations above capture the exact
semantics of Stalnaker-Lewis' counterfactuals but pose computational
and interpretability challenges.
DNF minimization may compress an extracted explanation but cannot
reduce the cost of closest-world enumeration.
We therefore consider a less general explanation that still captures
the counterfactual argument and guarantees prediction change.
We call it a \emph{conjunctive counterfactual explanation}, restricted
to conjunctions of literals rather than arbitrary formulas defined via
minimal Hamming distance.\footnote{%
  Conjunctive counterfactual explanations are not semantically
  equivalent to the closest-world explanations of the modal-logic
  account~\citep{lorini-logcom23}, but form a syntactic subset of
  counterfactual antecedents designed for better scalability and
  interpretability.
}
%
% In the rest of the paper, we will focus solely on conjunctive
% counterfactual explanations.

Such counterfactuals blend AXps and CXps: like CXps, they concern
prediction change, but CFXps \emph{guarantee}, i.e.\ logically entail,
it.
We use notation similar to that of abductive and contrastive
explanations, where an explanation need not be (cardinality- or)
subset-minimal.

\begin{definition} \label{def:cfxp}
  Let $\fml{E}=(M,\sbf{i})$ be an explanation problem with classifier
  $M=(\varphi,\mbb{F},\mbb{K},\kappa)$ and instance
  $\sbf{i}=(\sbf{v},c)$.
  A weak \emph{(conjunctive) counterfactual explanation} (weak CFXp)
  is a subset of features $\gamma \subseteq \varphi$ such that,
  \begin{equation} \label{eq:cfxp}
    \exists \delta\subseteq\varphi\setminus\gamma,
    \exists \sbf{y}\in[\sbf{v}]_{\varphi\setminus\gamma},
    \forall \sbf{x}\in[\sbf{y}]_{\gamma}\cap[\sbf{v}]_\delta.
    \quad \kappa(\sbf{x})\neq c \\
  \end{equation}
\end{definition}
Point $\sbf{y}$ is a \emph{witness} for $\gamma$: it is \emph{close
enough} to $\sbf{v}$ but predicted differently, and thus determines
the minimal changes from $\sbf{v}$.
Support $\delta$ fixes values shared by $\sbf{y}$ and $\sbf{v}$ so
that every point in $[\sbf{y}]_\gamma \cap [\sbf{v}]_\delta$ has a
prediction other than $c$, i.e.\ it \emph{``pins down what stays
fixed''}.
Thus, $\delta$ supports $\gamma$ and helps minimize it, capturing the
counterfactual conditional: $\gamma$ controls \emph{``variance''} from
$\sbf{v}$, while $\delta$ controls \emph{``invariance''}.
We write $(\gamma,\sbf{y},\delta)$ for a CFXp with its witness and
support.
Features in $\gamma$ take their values from $\sbf{y}$, features in
$\delta$ retain their values from $\sbf{v}$, and all others may vary
while the counterfactual prediction remains guaranteed.
(The case $\delta=\varphi\setminus\gamma$ is degenerate because
$[\sbf{y}]_{\gamma} \cap [\sbf{v}]_{\varphi\setminus\gamma} =
\{\sbf{y}\}$, making the explanation \emph{too narrow};
$\delta=\emptyset$ may also be degenerate when points
$[\sbf{y}]_{\gamma} \cap [\sbf{v}]_{\emptyset}$ are \emph{too far}
from $\sbf{v}$.
We show below that a ``good'' $\delta$ can be computed efficiently.)

As with AXps and CXps, a (strong conjunctive) counterfactual
explanation (CFXp) is a subset-minimal weak CFXp
$\gamma\subseteq\varphi$.
% %
% (In other words, $\gamma$ is a CFXp if it is subset-minimal.)
%
For CFXps, targeting a specific prediction may be more useful than
requiring \emph{any other} prediction.
For $q\in\mbb{K}$, $q\neq c$, a weak \emph{targeted counterfactual
explanation} (weak targeted CFXp) is a subset $\gamma$ such that,
\begin{equation} \label{eq:tcfxp}
  \exists \delta\subseteq\varphi\setminus\gamma,
  \exists \sbf{y}\in[\sbf{v}]_{\varphi\setminus\gamma},
  \forall \sbf{x}\in[\sbf{y}]_{\gamma}\cap[\sbf{v}]_\delta.
  \quad \kappa(\sbf{x})= q \\
\end{equation}
This specifies which features change, and how, to reach class
$q\neq c$.

% \begin{example} \label{ex:cfxp}
%   %
%   Consider the classification problem of \autoref{ex:xps} and a point
%   $\sbf{v}=(5.1,3.5,3.0,1.5)$, which is predicted as class
%   ``versicolor'' (recall that features 3 and 4 are \emph{petal.length}
%   and \emph{petal.width} respectively.).
%   %
%   A possible CFXp is $\gamma_1=\{3\}$.
%   %
%   According to the definition above, there is a witness
%   $\sbf{y}_1=(5.1,3.5,\ast 2.0\ast,1.5)$ predicted as ``setosa'';
%   furthermore, there is $\delta_1=\{4\}$ s.t.\ any point in
%   $[\sbf{y}_1]_{3}\cap[\sbf{v}]_{\{4\}}$ is guaranteed to be predicted
%   as ``setosa'' as the first rule in \autoref{fig:ds} will be bound to
%   fire the prediction.
%   %
%   Notably, $\gamma_1$ can be used to target either class ``setosa'' or
%   ``virginica''.
%   %
%   While the former has been already demonstrated, the latter can be
%   observed if one considers a witness $\sbf{y}_2=(5.1,3.5,\ast
%   5.0\ast,1.5)$: any point in $[\sbf{y}_2]_{3}\cap[\sbf{v}]_{\{4\}}$
%   is guaranteed to be predicted as ``virginica'' due to rule 4, using
%   the same supporting set $\delta_1=\{4\}$.
%   %
%   Finally, instance $\sbf{v}$ has another CFXp $\gamma_2=\{4\}$
%   supported by $\delta=\{3\}$ and the witness
%   $\sbf{y}_3=(5.1,3.5,3.0,\ast 2.0 \ast)$ as any point in
%   $[\sbf{y}_3]_{\{4\}}\cap[\sbf{v}]_{\{3\}}$ is classified as
%   ``virginica'', thanks to rule 3.
%   %
% \end{example}

\begin{example} \label{ex:toy1cfxp}
  For $\sbf{v}=(1,0,0)$, one CFXp is $\gamma_1=\{2\}$, witnessed by
  $\sbf{v}_7=(1,1,0)$ and supported by $\delta_1=\{1\}$, since every
  point in $[\sbf{v}_7]_{\{2\}} \cap [\sbf{v}]_{\{1\}}$ is classified
  as $1$.
  Thus, the triple $(\gamma_1,\sbf{v}_7,\delta_1) = (\{2\}, (1,1,0),
  \{1\})$ gives the counterfactual argument:
  \emph{``changing $x_2$ to $1$ while keeping $x_1$ fixed at $1$ will
  lead to prediction $1$, regardless of the value of $x_3$''}.
  Another is $\gamma_2=\{3\}$, witnessed by $\sbf{v}_6=(1,0,1)$ with
  support $\delta_2=\emptyset$, since every point in
  $[\sbf{v}_6]_{\{3\}}$ is classified as $1$.
  The triple $(\gamma_2, \sbf{v}_6, \delta_2) \triangleq
  (\{3\}, (1,0,1), \emptyset)$ gives the counterfactual argument:
  \emph{``changing $x_3$ to $1$ regardless of the values of $x_1$ and
  $x_2$ leads to prediction $1$''}.
\end{example}

% \begin{example} \label{ex:toy2cfxp}
%   %
%   Now, consider the explanation problem $\fml{E}_2$ from
%   \autoref{ex:toy2}, \autoref{tab:toy2func}, and
%   \autoref{ex:toy2dcfxp}.
%   %
%   Given the instance $\sbf{i}_2=(\sbf{u}, 0)$ with $\sbf{u}=(0,0,0)$,
%   a valid CFXp is $\gamma_1=\{1,2\}$,
%   witnessed by point $\sbf{u}_7=(1,1,0)$ and supported by a
%   subset of features $\delta_1=\emptyset$ because for any point
%   $[\sbf{u}_7]_{\{1,2\}} \cap [\sbf{u}]_{\emptyset}\triangleq
%   [\sbf{u}_7]_{\{1,2\}}$ (i.e.\ when $x_1=1 \land x_2=1$ holds), the
%   prediction is guaranteed to be $1$.
%   %
%   Thus, the triple $(\gamma_1,\sbf{u}_7,\delta_1) = (\{1,1\}, (1,1,0),
%   \emptyset)$ serves as a counterfactual argument:
%   %
%   \emph{``changing both $x_1$ and $x_2$ to $1$ will lead to prediction
%   $1$, regardless of the value of $x_3$''}.
%   %
%   Another CFXp is $\gamma_2=\{1,3\}$ witnessed by point
%   $\sbf{u}_{10}=(2,0,1)$ and supported by the subset of features
%   $\delta_2=\emptyset$.
%   %
%   Again, this is because any point in $[\sbf{u}_{10}]_{\{1,3\}} \cap
%   [\sbf{u}]_{\emptyset}$ (i.e. when $x_1=2 \land x_3=1$ holds) is
%   classified as $2$.
%   %
%   As a result, the triple $(\gamma_2, \sbf{u}_{10}, \delta_2)
%   \triangleq (\{1,3\}, (2,0,1), \emptyset)$ serves as a
%   counterfactual argument:
%   %
%   \emph{``changing $x_1$ to $2$ and $x_3$ to $1$ lead to prediction
%   $2$, regardless of the value of $x_2$''}.
%   %
%   \qed
%   %
% \end{example}

Both contrastive and counterfactual explanations suggest how to
change the prediction and are often called
\emph{actionable}~\citep{karimi-acmcs23,miller-corr24}.
Whereas CXps guarantee one nearby instance predicted differently from
$\sbf{v}$, CFXps may offer a family of differently predicted instances
based on $\gamma$.
In this sense, CFXps can be deemed stronger than CXps and viewed as
their generalization, as the following proposition shows:

\begin{proposition} \label{th:cxp}
  Let $\fml{E}=(M,\sbf{i})$ be an explanation problem with
  $M=(\varphi,\mbb{F},\mbb{K},\kappa)$ and
  $\sbf{i}=(\sbf{v},c)\in\mbb{F}\times\mbb{K}$.
  A subset $\gamma \subseteq \varphi$ is a weak targeted (untargeted,
  resp.) CFXp if and only if it is a weak targeted (untargeted, resp.)
  CXp.
  As a result, the subset-minimal feature sets of targeted
  (untargeted, resp.) CFXps and targeted (untargeted, resp.) CXps
  coincide.
\end{proposition}

\begin{proof}
  Let $\gamma$ be a weak \emph{untargeted} CFXp witnessed by point
  $\sbf{y} \in [\sbf{v}]_{\varphi \setminus \gamma}$ and supported by
  $\delta \subseteq \varphi \setminus \gamma$.
  Since $\sbf{y} \in [\sbf{y}]_\gamma \cap [\sbf{v}]_\delta$,
  \eqref{eq:cfxp} implies $\kappa(\sbf{y}) \neq c$.
  Hence, $\gamma$ is a weak CXp witnessed by $\sbf{y}$.
  Conversely, let $\gamma$ be a weak CXp witnessed by a point
  $\sbf{y} \in [\sbf{v}]_{\varphi \setminus \gamma}$ such that
  $\kappa(\sbf{y})\neq c$.
  Pick $\delta=\varphi\setminus\gamma$ and observe that
  $[\sbf{y}]_\gamma\cap[\sbf{v}]_\delta=\{\sbf{y}\}$.
  Hence, \eqref{eq:cfxp} holds and $\gamma$ is a weak CFXp.
  The targeted case follows similarly by using \eqref{eq:tcfxp} and
  the definition of targeted CXps~\citep{inams-aiia20}.
  Thus, the feature sets of weak CFXps and weak CXps coincide, as do
  their subset-minimal feature sets.
\end{proof}

\autoref{th:cxp} shows that CFXps and CXps have the same
subset-minimal feature sets.
A CFXp also associates $\gamma$ with a witness $\sbf{y}$ and support
$\delta$, guaranteeing the counterfactual prediction beyond $\sbf{y}$.

Because they capture entailment, \emph{untargeted} CFXps relate to the
negation of prime \emph{implicates} of the predicate $\kappa_c
\triangleq (\kappa(\sbf{v})=c)$.
Similarly, \emph{targeted} CFXps relate to \emph{absolute} (global or
class) abductive explanations~\citep{inms-nips19} (see also
\autoref{sec:comp}).
Given an instance $\sbf{i}=(\sbf{v},c)$, a CFXp $\gamma$ witnessed by
$(\sbf{y},q)$ and supported by $\delta$ guarantees the prediction
$q\neq c$.
Equivalently, $\gamma \cup \delta$ applied to $\sbf{y}$ is either a
counterexample for class $c \in \mbb{K}$ or an absolute explanation
for class $q\neq c$.
Next, we show how to compute targeted CFXps and relate them to AXps and
CXps.

\section{Computing Conjunctive Counterfactuals} \label{sec:comp}

We focus solely on \emph{targeted} CFXps, omitting ``targeted'' for
brevity.
We also focus on strong (subset-minimal) explanations.
%
% Hereinafter, we focus on strong (subset-minimal) explanations.
%
Given an instance of interest
$\sbf{i}=(\sbf{v},c)\in\mbb{F}\times\mbb{K}$, we show how a CXp
$\beta\subseteq\varphi$ for $\sbf{i}$ can be used to construct an
AXp $\alpha\subseteq\varphi$ for another instance $(\sbf{y},q)$,
$q\neq c$, witnessing $\beta$.

\begin{proposition} \label{th:nosub}
  Let $\beta\subseteq\varphi$ be a CXp for instance
  $\sbf{i}=(\sbf{v},c)$.
  Also, let $(\sbf{y},q)$ be an arbitrary witness of $\beta$, i.e.\
  $\sbf{y}\in[\sbf{v}]_{\varphi\setminus\beta}$.
  Then there is no AXp $\alpha$ for instance $(\sbf{y},q)$ such that
  $\alpha\subsetneq\beta$.
\end{proposition}

\begin{proof}
  Let $\beta$ be a CXp for $\sbf{i}=(\sbf{v},c)$ with witness
  $\sbf{y}\in[\sbf{v}]_{\varphi\setminus\beta}$ satisfying
  $\kappa(\sbf{y})=q$.
  Assume, by contradiction, that instance $(\sbf{y},q)$ has an AXp
  $\alpha \subsetneq \beta$.
  By~\eqref{eq:axp}, $\forall \sbf{x}\in[\sbf{y}]_\alpha.~
  \kappa(\sbf{x})=q$.
  Construct a point $\sbf{u}\in[\sbf{y}]_\alpha \cap
  [\sbf{v}]_{\varphi\setminus\alpha}$ by setting $u_i=y_i$ for
  $i\in\alpha$ and $u_j=v_j$ for $j\in\varphi\setminus\alpha$.
  Then $\sbf{u}\in[\sbf{y}]_\alpha$, and so $\kappa(\sbf{u})=q$.
  Yet $\sbf{u}$ differs from $\sbf{v}$ only on features
  $\alpha \subsetneq \beta$, contradicting that $\beta$ is a CXp.
\end{proof}

\begin{proposition} \label{th:sup}
  Let $\beta\subseteq\varphi$ be a CXp for instance
  $\sbf{i}=(\sbf{v},c)$ with an arbitrary witness $(\sbf{y},q)$, where
  $\sbf{y}\in[\sbf{v}]_{\varphi\setminus\beta}$ and
  $\kappa(\sbf{y})=q\neq c$.
  Then there is an AXp $\alpha$ for instance $(\sbf{y},q)$ such that
  $\alpha\supseteq\beta$.
\end{proposition}

\begin{proof}
  Let $\beta\subseteq\varphi$ be a CXp for instance
  $\sbf{i}=(\sbf{v},c)$ and fix a witness
  $\sbf{y}\in[\sbf{v}]_{\varphi\setminus\beta}$ with
  $\kappa(\sbf{y})=q$.
  Consider the following set of feature subsets:
  \begin{equation*}
    \Zeta = \{\zeta\subseteq\varphi\mid \beta \subseteq \zeta,
    \forall \sbf{x} \in [\sbf{y}]_\zeta.~ \kappa(\sbf{x})=q\}.
  \end{equation*}
  Thus, $\Zeta$ is the set of all weak AXps of $(\sbf{y},q)$ that
  include $\beta$.
  It is non-empty because $\varphi\in\Zeta$, as
  $[\sbf{y}]_\varphi=\{\sbf{y}\}$.
  Since $\Zeta$ is finite, it has a minimal element $\alpha\in\Zeta$
  wrt.\ \emph{set inclusion}.

  By construction, $\alpha$ satisfies three conditions:
  (1)~$\beta\subseteq\alpha$,
  (2)~$\forall \sbf{x}\in[\sbf{y}]_\alpha.~ \kappa(\sbf{x})=q$, i.e.\
  $\alpha$ is a weak AXp for $(\sbf{y},q)$, and
  (3)~minimality within $\Zeta$.

  % %
  % \begin{enumerate}
  %   \setlength{\itemsep}{-4pt}
  %   %
  %   \item $\beta\subseteq\alpha$;
  %   %
  %   \item $\forall \sbf{x}\in[\sbf{y}]_\alpha.~ \kappa(x)=q$, i.e.\
  %     $\alpha$ is a weak AXp for $(\sbf{y},q)$;
  %   %
  %   \item minimality within $\Zeta$.
  %   %
  % \end{enumerate}

  It remains to show that $\alpha$ is \emph{globally} subset-minimal,
  not only among weak AXps containing $\beta$.
  Suppose, for contradiction, that there exists $\alpha' \subsetneq
  \alpha$ such that $\forall \sbf{x} \in [\sbf{y}]_{\alpha'}.~
  \kappa(\sbf{x})=q$.
  Then necessarily $\alpha' \not\supseteq \beta$; otherwise, the
  minimality of $\alpha$ in $\Zeta$ would be contradicted.

  % It remains to show that $\alpha$ is \emph{globally} subset-minimal,
  % i.e.\ that it is an AXp.
  % %
  % Suppose, for contradiction, that there exists $\alpha' \subsetneq
  % \alpha$ such that $\forall \sbf{x} \in [\sbf{y}]_{\alpha'}.~
  % \kappa(\sbf{x})=q$.
  % %
  % Then necessarily $\alpha' \not\supseteq \beta$, or else the
  % minimality of $\alpha$ in $\Zeta$ would be contradicted.

  Construct a point $\sbf{u}$ by fixing all features in $\alpha'$ to
  $\sbf{y}$'s values and all other features to $\sbf{v}$'s values:
  \begin{equation*}
    u_j \;=\;
    \begin{cases}
      y_j,& j\in \alpha' \\
      v_j,& j\notin \alpha'
    \end{cases}
    \qquad\Longrightarrow\qquad
    \sbf{u} \in [\sbf{y}]_{\alpha'}\cap [\sbf{v}]_{\varphi\setminus
    \alpha'}.
  \end{equation*}
  Since $\alpha'$ is, by assumption, a weak AXp for $(\sbf{y},q)$ and
  $\sbf{u}\in[\sbf{y}]_{\alpha'}$, we have $\kappa(\sbf{u})=q$.

  Recall that $\sbf{y}$ is a witness for the CXp $\beta$, so we have
  $\{j \in \varphi \mid y_j\neq v_j\}=\beta$.
  Given that $u_j=v_j$ for all $j \not\in \alpha'$, the coordinates on
  which $\sbf{u}$ and $\sbf{v}$ \emph{actually} differ are exactly:
  \begin{equation*}
  \{j \in \varphi \mid u_j \neq v_j\}= \{j \in \alpha' \mid y_j \neq
  v_j\} = \alpha' \cap \beta \subseteq \beta.
  \end{equation*}
  Since $\alpha' \not\supseteq \beta$, there is a feature $i\in
  \beta\setminus \alpha'$, and so $\alpha' \cap \beta \subsetneq
  \beta$.
  Thus, we have produced a point $\sbf{u}$ with $\kappa(\sbf{u})=q$
  that differs from $\sbf{v}$ only on features $\alpha'\cap \beta
  \subsetneq \beta$.
  This contradicts that  $\beta$ is a CXp.
\end{proof}

\autoref{th:sup} has two important consequences.
First, it yields an efficient way to compute an AXp for an instance
$(\sbf{y},q)$ once this instance is known to witness a CXp $\beta$ for
$(\sbf{v},c)$.
The AXp computation can bypass traversing the entire feature set
$\varphi$, focusing only on features $\varphi \setminus \beta$.
This is useful when multiple instances from different classes are
explained in sequence, both in terms of AXps and CXps.
%
% First, once $(\sbf{y},q)$ witnesses a CXp $\beta$ for $(\sbf{v},c)$,
% an AXp for $(\sbf{y},q)$ can be computed by testing only
% $\varphi\setminus\beta$, rather than the complete feature set.
%
% This is useful when explaining multiple instances: CXps and witnesses
% computed for one instance can be cached and reused to compute AXps for
% witness instances.
%
It may help when compiling an ML model into a decision set or
enumerating explanations~\citep{inams-aiia20} for multiple instances.

Second, \autoref{th:sup} provides the basis for computing CFXps:

\begin{proposition} \label{th:cfxp}
  Let $\fml{E}=(M,\sbf{i})$ be an explanation problem with
  $M=(\varphi,\mbb{F},\mbb{K},\kappa)$ and
  $\sbf{i}=(\sbf{v},c)\in\mbb{F}\times\mbb{K}$.
  Let $\beta\subseteq\varphi$ be a CXp for $\sbf{i}$ witnessed by
  $(\sbf{y},q)$, and let $\alpha \supseteq \beta$ be an AXp for
  $(\sbf{y},q)$.
  %
  % Then $\beta$ is a CFXp for instance $\sbf{i}$ witnessed by $\sbf{y}$,
  % with a supporting set $\delta = \alpha \setminus \beta$.
  %
  Then $\beta$ is a CFXp for $\sbf{i}$, witnessed by $\sbf{y}$ and
  supported by $\delta = \alpha \setminus \beta$.
\end{proposition}

\begin{proof}
  We first show that $\beta$ is a weak CFXp satisfying
  \eqref{eq:tcfxp}.
  Since $\alpha$ is an AXp for $(\sbf{y},q)$, $\forall \sbf{x} \in
  [\sbf{y}]_\alpha.~ \kappa(\sbf{x})=q$.
  Since $\sbf{y}$ witnesses $\beta$, it agrees with $\sbf{v}$ on
  $\varphi\setminus\beta$, and hence on
  $\delta=\alpha\setminus\beta$.
  Thus, $[\sbf{v}]_\delta=[\sbf{y}]_\delta$.
  Hence, $[\sbf{y}]_\beta \cap [\sbf{v}]_\delta = [\sbf{y}]_\beta \cap
  [\sbf{y}]_\delta$. By \autoref{th:lemma}, $[\sbf{y}]_\beta \cap
  [\sbf{y}]_\delta = [\sbf{y}]_{\beta \cup \delta} =
  [\sbf{y}]_\alpha$.
  Therefore, $\forall \sbf{x}\in[\sbf{y}]_\beta \cap
  [\sbf{v}]_\delta.~\kappa(\sbf{x})=q$, so \eqref{eq:tcfxp} holds and
  $\beta$ is a weak CFXp.

  It remains to show that $\beta$ is subset-minimal.
  Otherwise, some $\beta' \subsetneq \beta$ is a weak CFXp.
  Then, by \autoref{th:cxp}, $\beta'$ is a weak CXp, contradicting
  that $\beta$ is a CXp.
  Hence, $\beta$ is a CFXp.
\end{proof}

By \autoref{th:cfxp}, any CXp for $\sbf{i}=(\sbf{v},c)$ can be
declared a CFXp by computing the witness and support as follows:
\rev{
  (1)~find any CXp $\beta$ for instance $\sbf{i}$ with witness
  $(\sbf{y},q)$,
  (2)~compute any AXp $\alpha\supseteq \beta$ for $(\sbf{y},q)$, and
  (3)~output a triple $(\beta, \sbf{y}, \alpha \setminus \beta)$.
}
%
% \begin{enumerate}
%   %
%   \setlength{\itemsep}{-2pt}
%   %
%   \item find any CXp $\beta$ for instance $\sbf{i}$ with witness
%     $(\sbf{y},q)$;
%     %
%   \item compute any AXp $\alpha\supseteq \beta$ for $(\sbf{y},q)$;
%     %
%   \item output a triple $(\beta, \sbf{y}, \alpha \setminus \beta)$.
%   %
% \end{enumerate}
%
The triple $(\beta,\sbf{y},\alpha\setminus\beta)$ gives the CFXp, its
witness, and its support, respectively.
The support need not be unique: different traversal orders may yield
different subset-minimal supports without affecting correctness.
\rev{\autoref{alg:cfxp} details this procedure with each test being a
reasoning oracle call checking the validity of \eqref{eq:cxp} and then
\eqref{eq:axp}.}

\rev{
  \autoref{th:sup} and \autoref{th:cfxp} hold for \emph{any} witness
  of $\beta$.
  %
  % This means the witness may optionally be improved (see
  % \autoref{ln:opt}).
  %
  This means we may optionally consider optimizing the
  \emph{quality} of the witness being computed (see
  \autoref{ln:opt}), if we are not satisfied with the \emph{default}
  one.
  %
  % The optional \textsc{Refine} step must preserve the coordinates
  % outside $\beta$ and return a witness with prediction different
  % from $c$.
  %
  For instance, when a winning class $q\in\mbb{K}$ is determined by
  its \emph{score} $s_q(\sbf{y})$, witnesses $\sbf{y}$ lying near the
  decision boundary, i.e.\ when $s_q(\sbf{y})-\max_{k\neq
  q}s_k(\sbf{y})$ is close to 0, may lead to a large support set
  $\delta$.
  To avoid that, we can maximize this \emph{margin} of class $q$
  subject to the CXp $\beta$.
  For ReLU networks, this can be done using one MILP call.
  (Minimizing the magnitude of the change can be done similarly.)
  Finally, since every CXp yields a CFXp, the enumeration of
  CXps~\citep{inams-aiia20} directly yields different CFXps with
  pairwise different changed features.
}

\begin{revblock}
\renewcommand{\normalcolor}{\color{revcolor}}
\begin{algorithm2e}[t]
  \color{revcolor}
  \SetKw{KwTo}{in}\SetKwFor{For}{for}{\string:}{}%
  \SetKwIF{If}{ElseIf}{Else}{if}{:}{elif}{else:}{}%
  \SetKwFor{While}{while}{:}{}%
  \SetKwFor{ForEach}{foreach}{:}{}%
  \AlgoDontDisplayBlockMarkers\SetAlgoNoEnd\SetAlgoNoLine%
  \LinesNumbered
  \DontPrintSemicolon%
  \SetKw{KwNot}{not\xspace}
  \SetKw{KwAnd}{and\xspace}
  \SetKw{KwOr}{or\xspace}
  \SetKw{KwBreak}{break\xspace}
  \SetKwData{false}{{\small false}}
  \SetKwData{true}{{\small true}}
  \SetKwData{st}{{\small \slshape st}}
  \SetKwData{cost}{{\small \slshape cost}}
  \SetKwFunction{CNF}{CNF}
  \SetKwFunction{SAT}{SAT}
  \SetKwFunction{refine}{Refine}
  \SetKwInOut{Input}{input}\SetKwInOut{Output}{output}
  \newcommand\commfont[1]{\footnotesize\em{#1}}
  \SetKwComment{tcpy}{\# }{}
  \SetCommentSty{commfont}

  \footnotesize
  \DontPrintSemicolon
  \SetKwInOut{Input}{Input}
  \SetKwInOut{Output}{Output}
  \Input{Classifier $\kappa$, instance $\sbf{i}=(\sbf{v},c)$}
  \Output{CFXp $\beta$, witness $\sbf{y}$, support $\delta$}
  $(\alpha,\beta)\gets(\varphi,\varphi)$; ~~$\sbf{y}\gets$ a point with
  $\kappa(\sbf{y})\neq c$ \tcpy*[r]{initialization; $\sbf{y}$ exists unless $\kappa$ is constant}
  \ForEach(\tcp*[f]{CXp extraction}){$j\in\varphi$}{
    \lIf{$\exists\sbf{x}\in[\sbf{v}]_{\varphi\setminus\beta\cup\{j\}}.~\kappa(\sbf{x})\neq c$}{
      $\beta\gets\beta\setminus\{j\}$; $\sbf{y}\gets\sbf{x}$
    }
  }
  $\sbf{y}\gets\refine(\sbf{y}, \kappa, c, \beta)$\label{ln:opt}\tcpy*[4]{\textbf{optional}; improve $\sbf{y}$ changing only features in $\beta$, with $\kappa(\sbf{y})\neq c$}
  $q\gets\kappa(\sbf{y})$\tcpy*[r]{witness $\sbf{y}$ is classified as class $q$}
  \ForEach(\tcp*[f]{AXp extraction, \autoref{th:sup}}){$i\in\varphi\setminus\beta$}{
    \lIf{$\forall\sbf{x}\in[\sbf{y}]_{\alpha\setminus\{i\}}.~\kappa(\sbf{x})=q$}{
      $\alpha\gets\alpha\setminus\{i\}$
    }
  }
  \Return{$(\beta,\sbf{y},\alpha\setminus\beta)$}\tcpy*[r]{returning the complete triple}
  \caption{Computing a CFXp}
  \label{alg:cfxp}
\end{algorithm2e}
\end{revblock}
\renewcommand{\normalcolor}{\color{black}}

This construction neither uses a distance function nor enumerates
closest counterexamples: it computes a CXp and witness followed by an
AXp for that witness.
With deletion-based extraction, both steps make linearly many reasoner
calls; once $\beta$ and its witness are known, the second tests at most
$|\varphi|-|\beta|$ features.
These are call bounds only; each call's cost depends on the model
representation and reasoning engine.
Closest-world extraction is also output-sensitive because the number
of closest counterexamples may be exponential.

\begin{example} \label{ex:toy1cfcomp}
  In the running example, $\gamma_1=\{2\}$ and $\gamma_2=\{3\}$ are
  obtained from CXps $\beta_1=\{2\}$ and $\beta_2=\{3\}$ by computing
  AXps for witnesses $\sbf{v}_7$ and $\sbf{v}_6$.
  From $\sbf{v}_7$ we obtain $\alpha_1=\{1,2\}$ and the triple
  $(\gamma_1,\sbf{v}_7,\{1\})$ by testing features $1$ and $3$ but,
  by \autoref{th:sup}, not feature $2\in\beta_1$.
  For $\beta_2$, we analogously obtain
  $(\gamma_2,\sbf{v}_6,\emptyset)$.
\end{example}

The \emph{AXp-CXp-CFXp triad} holds only for conjunctive CFXps, not
for closest-world counterfactuals~\citep{lorini-logcom23}: CFXps
\emph{relax} the exhaustive universal condition on minimal-distance
instances~\citep{stalnaker-theory68,lewis-counterfactuals73} into the
more actionable existential--universal pattern \eqref{eq:tcfxp}.

\section{Experimental Evidence} \label{sec:res}

This section presents experimental results assessing the performance
of a single CFXp extraction on widely studied public datasets.
First, we compare CFXps with closest-world counterfactual explanations
(CWCs), which directly instantiate the closest-world semantics
formalized in \autoref{sec:fcc}.
Second, we compare CFXps with well-established point-based
counterfactual explainers on XGBoost tree ensembles and ReLU neural
networks.

\paragraph{Experimental setup.} \label{sec:setup}
The experiments were conducted on a MacBook Pro with an Apple M1 Pro
CPU with 10 cores and 32\,GByte RAM running macOS 26.2 Tahoe.
The time limit for extracting an individual formal explanation is set
to 5 minutes; no memory limit is imposed.
We use 21 tabular datasets from the UCI Machine Learning
Repository~\citep{kelly-uci23} and Penn Machine Learning Benchmarks
(PMLB)~\citep{olson-biodm17,romano-corr21}.
The number of original features ranges from 7 to 60 and the number of
instances from 59 to 58000, with averages of 26.95 and 4684.47,
respectively.
For each dataset, we use the same randomly selected set of at most 200
instances, resulting in 3755 explained predictions.

\paragraph{Implementations.} \label{sec:prototype}
For tree ensembles, we reuse the XGBoost models and experimental setup
of the earlier MaxSAT study~\citep{iisms-aaai22}.
The models contain 50 trees per class of depth at most 3--4, and their
propositional encodings use 4--10183 variables and 304--102611
clauses.
The MaxSAT approach~\citep{iisms-aaai22} encodes tree-induced
intervals and paths as hard clauses and retains floating-point leaf
scores as weights of soft objective literals; feature relevance is
tested by incrementally invoking the RC2 MaxSAT
solver~\citep{imms-sat18,imms-jsat19}.
For neural networks, we train a fully connected ReLU model on each of
the same 21 datasets, with hidden layers of 16 and 8 neurons.
%
% Every input column is treated as numerical and scaled to $[0,1]$; the
% models use the same seed-7 80/20 split as XGBoost.
%
The resulting models have 7--60 inputs, 2--11 classes, and 273--1121
trainable parameters.
They are encoded into MILP using a standard ReLU
encoding~\citep{fischetti-cj18} and solved with Gurobi~\citep{gurobi}.
Thus, the experiments use two independent reasoning backends for CXp,
AXp, and CFXp extraction.

\paragraph{Comparison with closest-world explanations.}
\label{sec:cwc-res}
This experiment is restricted to tree ensembles, whose thresholds
partition every continuous feature domain into a finite number of
non-overlapping intervals.
As a result, closest worlds can be enumerated over the corresponding
regions.
To this end, we extend the MaxSAT approach for tree
ensembles~\citep{iisms-aaai22} to enumerate all smallest-size CXps and
all their witnesses.
In the following, this procedure is referred to as CWC and is compared
with the computation of one AXp, CXp, and CFXp per prediction.

CFXp computation remains comparable to AXp and CXp computation, as it
consists of computing a CXp followed by an AXp.
All three methods finish on all 3755 instances, with mean/median
runtimes of 1.529/0.266, 0.783/0.184, and 2.098/0.500 seconds for AXp,
CXp, and CFXp extraction, respectively.
By contrast, CWC reaches the 300-second time limit in 428 cases.
Keeping these timeout observations capped at 300 seconds gives a
mean/median runtime of 51.872/2.168 seconds.
The true mean is necessarily higher because these 428 runs did not
finish within the time limit.

The higher computational cost of CWC should be attributed to the
number of closest worlds that must be enumerated before the
computation can finish.
For the 3327 completed cases, this number ranges from 1 to 17697, with
mean/median 117.770/19.
Regarding explanation size, AXps have mean/median size 9.043/8, while
the CFXp feature set $\gamma$, its support $\delta$, and the complete
certificate have mean/median sizes 1.972/1, 8.240/7, and 10.212/8,
respectively.
Although the number of worlds and the number of features measure
different objects, these are nevertheless the natural sizes of the two
representations.
The complete runtime and size distributions, including cactus and
per-instance scatter plots, are provided in the appendix.

\paragraph{Comparison with point-based explainers.}
\label{sec:point-res}
We next compare CFXps with DiCE~\citep{mothilal-fat20}, Growing
Spheres~\citep{laugel-corr17}, and the optimization-based method of
Wachter et al.~\citep{wachter-hjlt18}.
Each method is asked for one point whose prediction differs from that
of the explained instance.
All returned points are checked independently against the
corresponding model; hence, the success counts in
\autoref{tab:point-res} include only model-verified points.
For multiclass problems, CFXps and Growing Spheres are untargeted,
whereas DiCE and Wachter require a target class.
Therefore, we run the latter for every class other than the original
prediction and select the point changing the fewest features.
All three methods are evaluated on the same 3755 predictions from all
21 datasets.
%
% Following the XReason protocol, every original input column is
% treated as numerical.

\begin{table}[!ht]
  \small
  \caption{Comparison with point-based counterfactual explainers. Size
  and time entries report \emph{mean/median}. $L_0$ distance denotes the
  number of features whose values differ from the original instance.
  These and paired CFXp $|\gamma|$ are computed over
  model-verified point outputs;
  CFXp$\leq$ reports the fraction of these pairs in which $|\gamma|$
  is no larger. Times cover all attempted instances and use CPU time
  for XGBoost and wall time for ReLU networks.}
  \label{tab:point-res}
  \begin{adjustbox}{max width=\textwidth,center}
  \setlength{\tabcolsep}{3.5pt}
  \begin{tabular}{llrrrrrr}
    \toprule
    Model & Point method & Verified & $L_0$ distance &
    CFXp $|\gamma|$ & CFXp$\leq$ & Point time (s) & CFXp time (s) \\
    \midrule
    \multirow{3}{*}{XGBoost}
      & DiCE             & 3754/3755 & 2.310/{\color{white}0}2 & 1.971/1 & 87.24\%  & 0.483/{\color{white}0}0.208 & 2.098/0.500 \\
      & Growing Spheres  & 3755/3755 & 3.512/{\color{white}0}3 & 1.972/1 & 95.31\%  & 5.518/{\color{white}0}3.428 & 2.098/0.500 \\
      & Wachter          & 3461/3755 & 20.588/16               & 1.969/1 & 100.00\% & 41.128/23.985               & 2.098/0.500 \\
    \midrule
    \multirow{3}{*}{ReLU}
      & DiCE             & 3755/3755 & 2.083/{\color{white}0}2 & 2.064/1 & 79.92\%  & 0.408/{\color{white}0}0.196 & 0.310/0.076 \\
      & Growing Spheres  & 3692/3755 & 9.836/{\color{white}0}9 & 1.882/1 & 97.24\%  & 27.550/11.325               & 0.310/0.076 \\
      & Wachter          & 3357/3755 & 20.120/15               & 1.982/1 & 99.73\%  & 4.163/{\color{white}0}2.588 & 0.310/0.076 \\
    \bottomrule
  \end{tabular}
  \end{adjustbox}
\end{table}

As shown in \autoref{tab:point-res}, DiCE, Wachter, and Growing
Spheres return a model-verified result in 99.97--100\%,
89.40--92.17\%, and 98.3--100\% of all cases.
CFXp extraction finishes on all 3755 predictions for both model
classes.
For the neural models, $|\gamma|$ ranges from 1 to 25 with mean/median
2.064/1, while extraction takes from 0.005 to 27.749 seconds with
mean/median 0.310/0.076 seconds.
The support $\delta$ and complete \emph{certificate}
$\gamma\cup\delta$ have mean/median sizes 20.577/17 and 22.641/19,
respectively.
For context, ordinary AXps for these models have mean/median size
19.734/15, which helps explain the larger support sets.
\rev{A \emph{optionally improved} witness reduces $\delta$ and
$\gamma\cup\delta$ to 19.312/16 and 21.375/18, at 4\,ms per instance
on average.}
Importantly, these support features remain fixed rather than being
changed.
Furthermore, for neural models the CFXp changed-feature set $\gamma$
is no larger than the number of changed features in the explanation of
DiCE, Wachter, and Growing Spheres, in 79.92\%, 99.73\%, and 97.24\%
of the paired cases.
For XGBoost models, the CFXp changed-feature set $\gamma$ is no larger
in 87.24\%, 100\%, and 95.31\% of the paired cases.
These results should not be interpreted as a general ranking of the
methods: in particular, the Wachter method optimizes normalized $L_1$
distance instead of $L_0$, while DiCE is faster than the MaxSAT-based
CFXp implementation on XGBoost.
Nevertheless, the results show that the strong regional guarantee
provided by a CFXp generally does not require more changed features
than a single model-verified point.
%
% \rev{Per-dataset results and bootstrap confidence intervals are given in
% \autoref{sec:app-robust}.}

\begin{revblock}

\paragraph{On CFXp regions.} \label{sec:extent-res}
Features outside $\gamma\cup\delta$ are \emph{free} to take any value.
For tree ensembles, 58.3\% of the features are free on average, and
only 0.2\% of the CFXps cover the witness alone.
For ReLU networks, these numbers are 17.4\% and 21.8\%.
% (as 35.1\% of
% the witnesses returned by MILP lie near the decision boundary).
%
Improving the witness raises the former percentage to 22.2\% and
lowers the latter to 11.8\%.

\paragraph{Strengthened point counterfactuals.}
\label{sec:point-axp-res}
An AXp can be computed for any instance, including a counterfactual
point returned by a point-based explainer.
Doing so generalizes such a counterfactual by providing a feature
region predicted as the point itself.
Unlike CFXps (thanks to \autoref{th:sup}), this AXp need not contain
the suggested changed features,
% i.e.\ the resulting subspace is not guaranteed to contain the
% suggested subset-minimal change,
which limits the practicality of such AXp-strengthened point
counterfactuals.
\autoref{tab:point-axp} compares CFXps with AXps obtained for all
model-verified points of \autoref{tab:point-res}.
DiCE's AXps produce regions of a similar extent but omit at least one
changed feature for 22.6\% (for XGBoost) and 9.2\% (for ReLU) of
points; among those they drop 43.0\% and 38.3\% of changed features,
respectively.
Growing Spheres yields smaller regions, while Wachter's regions are
not change-consistent in the lion's share of cases.
%
% Thus, AXps extend point counterfactuals to regions, but only CFXps
% guarantee a region containing a subset-minimal change.

\begin{table}[t]
  \color{revcolor}
  \footnotesize
  \captionsetup{labelfont={color=revcolor}}
  \caption{\color{revcolor} Point counterfactuals extended with AXps
  vs.\ CFXps (for ReLU models, using optionally improved witnesses).
  The columns are: mean free features, cases with no free feature,
  AXps omitting a changed feature (with mean proportion of omitted
  features), and mean/median time.}

  \label{tab:point-axp}
  \begin{adjustbox}{max width=\textwidth,center}
  \setlength{\tabcolsep}{5.5pt}
  \begin{tabular}{llrrrr}
    \toprule
    Model & Method & Free & Only point & Omits changed & Time (s) \\
    \midrule
    \multirow{4}{*}{XGBoost}
      & CFXp                  & 58.3\% & 0.2\%  & ---             & 2.098/0.500   \\
      & DiCE + AXp            & 57.3\% & 0.1\%  & 22.6\% (43.0\%) & 1.804/0.522   \\
      & Growing Spheres + AXp & 51.7\% & 1.0\%  & 0.7\% (24.0\%)  & 6.863/4.180   \\
      & Wachter + AXp         & 56.4\% & 0.2\%  & 99.2\% (49.9\%) & 45.122/25.437 \\
    \midrule
    \multirow{4}{*}{ReLU}
      & CFXp                  & 22.2\% & 11.8\% & ---             & 0.393/0.095   \\
      & DiCE + AXp            & 22.1\% & 11.8\% & 9.2\% (38.3\%)  & 0.663/0.321   \\
      & Growing Spheres + AXp & 8.6\%  & 37.8\% & 6.1\% (10.2\%)  & 22.987/11.092 \\
      & Wachter + AXp         & 19.7\% & 7.5\%  & 82.6\% (18.7\%) & 4.607/2.960   \\
    \bottomrule
  \end{tabular}
  \end{adjustbox}
\end{table}
\end{revblock}

\paragraph{Summary.} \label{sec:summary}
The experiments show that CFXp extraction is practically efficient for
tree ensembles and ReLU neural networks using two independent
reasoning backends.
Unlike CWC, CFXps require neither Hamming distance nor exhaustive
closest-world enumeration, and are not restricted to finite feature
domains.
Compared with popular point-based explainers, CFXps generally require
no more feature changes while guaranteeing the counterfactual
prediction beyond one returned instance.

\section{Conclusions} \label{sec:conc}

This work has developed a formal framework for counterfactual
explanations in classifier systems that connects closest-world
counterfactuals~\citep{stalnaker-theory68,lewis-counterfactuals73,lorini-logcom23}
with feature-based explanations used in formal XAI~\citep{msi-aaai22}.
We instantiated the modal-logic framework of~\citep{lorini-logcom23}
in a feature-space setting, obtaining disjunctive closest-world
counterfactual explanations under the Hamming-style similarity
measure.
We then introduced conjunctive counterfactual explanations (CFXps), a
restricted yet actionable class of formal feature-based
counterfactuals that guarantee prediction change and align naturally
with abductive explanations (AXps) and contrastive explanations
(CXps).
Unlike closest-world counterfactuals, CFXps require no distance
function and are not restricted to finite feature domains.

% On the theoretical side,
%
First,
we established a triad linking AXps, CXps, and CFXps.
Feature-wise, every CFXp is a CXp, and every CXp together with an AXp
for its witness instance induces a CFXp whose support set is
characterized by that AXp, yielding a modular algorithm that reuses
existing AXp/CXp machinery.
%
% On the practical side,
%
Second,
our experiments on XGBoost tree ensembles and ReLU neural networks
show that CFXp extraction is practically efficient.
% using independent MaxSAT and MILP backends.
%
For tree ensembles, CFXp computation remains comparable to single AXps
or CXps, while closest-world counterfactual explanations require
costly exhaustive enumeration.
Compared with popular point-based explainers, CFXps generally require
no more feature changes while providing a guarantee beyond one
returned point.

Overall, CFXps offer a viable compromise between the full
closest-world semantics of modal counterfactuals and the combinatorial
structure of formal AXps and CXps, providing sound, efficiently
computable, and comparatively succinct counterfactual arguments within
the formal XAI toolbox.

% \clearpage
\bibliographystyle{iclr2027_conference}
\bibliography{refs}

@article{rivest-ml87,
  author    = {Ronald L. Rivest},
  title     = {Learning Decision Lists},
  journal   = {Mach. Learn.},
  volume    = {2},
  number    = {3},
  pages     = {229--246},
  year      = {1987},
}

@inproceedings{clark-ewsl91,
  author    = {Peter Clark and
               Robin Boswell},
  title     = {Rule Induction with {CN2:} Some Recent Improvements},
  booktitle = {{EWSL}},
  pages     = {151--163},
  year      = {1991}
}

@article{lipton-cacm18,
  author    = {Zachary C. Lipton},
  title     = {The mythos of model interpretability},
  journal   = {Commun. {ACM}},
  volume    = {61},
  number    = {10},
  pages     = {36--43},
  year      = {2018},
  OPTurl       = {https://doi.org/10.1145/3233231},
  OPTdoi       = {10.1145/3233231},
  bibsource = {dblp computer science bibliography, https://dblp.org}
}

@article{marquis-corr21,
  author    = {Gilles Audemard and
               Steve Bellart and
               Louenas Bounia and
               Fr{\'{e}}d{\'{e}}ric Koriche and
               Jean{-}Marie Lagniez and
               Pierre Marquis},
  title     = {On the Computational Intelligibility of Boolean Classifiers},
  journal   = {CoRR},
  volume    = {abs/2104.06172},
  year      = {2021},
  url       = {https://arxiv.org/abs/2104.06172},
  archivePrefix = {arXiv},
  eprint    = {2104.06172},
  bibsource = {dblp computer science bibliography, https://dblp.org}
}

@inproceedings{hiims-kr21,
  author    = {Xuanxiang Huang and Yacine Izza and Alexey Ignatiev and Joao Marques{-}Silva},
  title     = {On Efficiently Explaining Graph-Based Classifiers},
  booktitle = {{KR}},
  OPTpages     = {},
  month     = {November},
  year      = {2021},
  OPTnote      = {Preprint available from \url{https://arxiv.org/abs/2106.01350}}
}

@inproceedings{msgcin-icml21,
  author    = {Joao Marques{-}Silva and
               Thomas Gerspacher and
               Martin C. Cooper and
               Alexey Ignatiev and
               Nina Narodytska},
  OPTeditor    = {Marina Meila and
               Tong Zhang},
  title     = {Explanations for Monotonic Classifiers},
  OPTbooktitle = {Proceedings of the 38th International Conference on Machine Learning,
               {ICML} 2021, 18-24 July 2021, Virtual Event},
  booktitle = {ICML},
  OPTseries    = {Proceedings of Machine Learning Research},
  OPTvolume    = {139},
  pages     = {7469--7479},
  OPTpublisher = {{PMLR}},
  year      = {2021},
  OPTurl       = {http://proceedings.mlr.press/v139/marques-silva21a.html},
  bibsource = {dblp computer science bibliography, https://dblp.org}
}

@inproceedings{inams-aiia20,
  author    = {Alexey Ignatiev and
               Nina Narodytska and
               Nicholas Asher and
               Joao Marques{-}Silva},
  OPTeditor    = {Matteo Baldoni and
               Stefania Bandini},
  title     = {From Contrastive to Abductive Explanations and Back Again},
  OPTbooktitle = {AIxIA 2020 - Advances in Artificial Intelligence - XIXth International
               Conference of the Italian Association for Artificial Intelligence,
               Virtual Event, November 25-27, 2020, Revised Selected Papers},
  booktitle = {AI*IA},
  OPTseries    = {Lecture Notes in Computer Science},
  OPTvolume    = {12414},
  pages     = {335--355},
  OPTpublisher = {Springer},
  year      = {2020},
  OPTurl       = {https://doi.org/10.1007/978-3-030-77091-4\_21},
  OPTdoi       = {10.1007/978-3-030-77091-4\_21},
  bibsource = {dblp computer science bibliography, https://dblp.org}
}

@article{iims-corr20,
  author    = {Yacine Izza and
               Alexey Ignatiev and
               Joao Marques{-}Silva},
  title     = {On Explaining Decision Trees},
  journal   = {CoRR},
  volume    = {abs/2010.11034},
  year      = {2020},
  url       = {https://arxiv.org/abs/2010.11034},
  archivePrefix = {arXiv},
  eprint    = {2010.11034},
  bibsource = {dblp computer science bibliography, https://dblp.org}
}

@inproceedings{barcelo-nips20,
  author    = {Pablo Barcel{\'{o}} and
               Mika{\"{e}}l Monet and
               Jorge P{\'{e}}rez and
               Bernardo Subercaseaux},
  OPTeditor    = {Hugo Larochelle and
               Marc'Aurelio Ranzato and
               Raia Hadsell and
               Maria{-}Florina Balcan and
               Hsuan{-}Tien Lin},
  title     = {Model Interpretability through the lens of Computational Complexity},
  OPTbooktitle = {Advances in Neural Information Processing Systems 33: Annual Conference
               on Neural Information Processing Systems 2020, NeurIPS 2020, December
               6-12, 2020, virtual},
  booktitle = {NeurIPS},
  year      = {2020},
  OPTurl       = {https://proceedings.neurips.cc/paper/2020/hash/b1adda14824f50ef24ff1c05bb66faf3-Abstract.html},
  bibsource = {dblp computer science bibliography, https://dblp.org}
}

@inproceedings{msgcin-nips20,
  author    = {Joao Marques{-}Silva and
               Thomas Gerspacher and
               Martin C. Cooper and
               Alexey Ignatiev and
               Nina Narodytska},
  OPTeditor    = {Hugo Larochelle and
               Marc'Aurelio Ranzato and
               Raia Hadsell and
               Maria{-}Florina Balcan and
               Hsuan{-}Tien Lin},
  title     = {Explaining Naive {Bayes} and Other Linear Classifiers with Polynomial
               Time and Delay},
  OPTbooktitle = {Advances in Neural Information Processing Systems 33: Annual Conference
               on Neural Information Processing Systems 2020, NeurIPS 2020, December
               6-12, 2020, virtual},
  booktitle = {NeurIPS},
  year      = {2020},
  OPTurl       = {https://proceedings.neurips.cc/paper/2020/hash/eccd2a86bae4728b38627162ba297828-Abstract.html},
  bibsource = {dblp computer science bibliography, https://dblp.org}
}

@inproceedings{marquis-kr20,
  author    = {Gilles Audemard and
               Fr{\'{e}}d{\'{e}}ric Koriche and
               Pierre Marquis},
  OPTeditor    = {Diego Calvanese and
               Esra Erdem and
               Michael Thielscher},
  title     = {On Tractable {XAI} Queries based on Compiled Representations},
  OPTbooktitle = {Proceedings of the 17th International Conference on Principles of
               Knowledge Representation and Reasoning, {KR} 2020, Rhodes, Greece,
               September 12-18, 2020},
  booktitle = {KR},
  pages     = {838--849},
  year      = {2020},
  OPTurl       = {https://doi.org/10.24963/kr.2020/86},
  OPTdoi       = {10.24963/kr.2020/86},
  bibsource = {dblp computer science bibliography, https://dblp.org}
}

@inproceedings{darwiche-ecai20,
  author    = {Adnan Darwiche and
               Auguste Hirth},
  OPTeditor    = {Giuseppe De Giacomo and
               Alejandro Catal{\'{a}} and
               Bistra Dilkina and
               Michela Milano and
               Sen{\'{e}}n Barro and
               Alberto Bugar{\'{\i}}n and
               J{\'{e}}r{\^{o}}me Lang},
  title     = {On the Reasons Behind Decisions},
  OPTbooktitle = {{ECAI} 2020 - 24th European Conference on Artificial Intelligence,
               29 August-8 September 2020, Santiago de Compostela, Spain, August
               29 - September 8, 2020 - Including 10th Conference on Prestigious
               Applications of Artificial Intelligence {(PAIS} 2020)},
  booktitle = {ECAI},
  OPTseries    = {Frontiers in Artificial Intelligence and Applications},
  OPTvolume    = {325},
  pages     = {712--720},
  OPTpublisher = {{IOS} Press},
  year      = {2020},
  OPTurl       = {https://doi.org/10.3233/FAIA200158},
  OPTdoi       = {10.3233/FAIA200158},
  bibsource = {dblp computer science bibliography, https://dblp.org}
}

@inproceedings{inms-nips19,
  author    = {Alexey Ignatiev and
               Nina Narodytska and
               Joao Marques{-}Silva},
  OPTeditor    = {Hanna M. Wallach and
               Hugo Larochelle and
               Alina Beygelzimer and
               Florence d'Alch{\'{e}}{-}Buc and
               Emily B. Fox and
               Roman Garnett},
  title     = {On Relating Explanations and Adversarial Examples},
  OPTbooktitle = {Advances in Neural Information Processing Systems 32: Annual Conference
               on Neural Information Processing Systems 2019, NeurIPS 2019, December
               8-14, 2019, Vancouver, BC, Canada},
  booktitle = {NeurIPS},
  pages     = {15857--15867},
  year      = {2019},
  OPTurl       = {https://proceedings.neurips.cc/paper/2019/hash/7392ea4ca76ad2fb4c9c3b6a5c6e31e3-Abstract.html},
  bibsource = {dblp computer science bibliography, https://dblp.org}
}

@inproceedings{inms-aaai19,
  author    = {Alexey Ignatiev and
               Nina Narodytska and
               Joao Marques{-}Silva},
  title     = {Abduction-Based Explanations for Machine Learning Models},
  OPTbooktitle = {The Thirty-Third {AAAI} Conference on Artificial Intelligence, {AAAI}
               2019, The Thirty-First Innovative Applications of Artificial Intelligence
               Conference, {IAAI} 2019, The Ninth {AAAI} Symposium on Educational
               Advances in Artificial Intelligence, {EAAI} 2019, Honolulu, Hawaii,
               USA, January 27 - February 1, 2019},
  booktitle = {AAAI},
  pages     = {1511--1519},
  OPTpublisher = {{AAAI} Press},
  year      = {2019},
  OPTurl       = {https://doi.org/10.1609/aaai.v33i01.33011511},
  OPTdoi       = {10.1609/aaai.v33i01.33011511},
  bibsource = {dblp computer science bibliography, https://dblp.org}
}

@article{miller-aij19,
  author    = {Tim Miller},
  title     = {Explanation in artificial intelligence: Insights from the social sciences},
  journal   = {Artif. Intell.},
  volume    = {267},
  pages     = {1--38},
  year      = {2019},
  OPTurl       = {https://doi.org/10.1016/j.artint.2018.07.007},
  OPTdoi       = {10.1016/j.artint.2018.07.007},
  bibsource = {dblp computer science bibliography, https://dblp.org}
}

@inproceedings{darwiche-ijcai18,
  author    = {Andy Shih and
               Arthur Choi and
               Adnan Darwiche},
  OPTeditor    = {J{\'{e}}r{\^{o}}me Lang},
  title     = {A Symbolic Approach to Explaining {Bayesian} Network Classifiers},
  OPTbooktitle = {Proceedings of the Twenty-Seventh International Joint Conference on
               Artificial Intelligence, {IJCAI} 2018, July 13-19, 2018, Stockholm,
               Sweden},
  booktitle = {IJCAI},
  pages     = {5103--5111},
  OPTpublisher = {ijcai.org},
  year      = {2018},
  OPTurl       = {https://doi.org/10.24963/ijcai.2018/708},
  OPTdoi       = {10.24963/ijcai.2018/708},
  bibsource = {dblp computer science bibliography, https://dblp.org}
}

@inproceedings{mothilal-fat20,
  author       = {Ramaravind Kommiya Mothilal and
                  Amit Sharma and
                  Chenhao Tan},
  title        = {Explaining machine learning classifiers through diverse counterfactual
                  explanations},
  booktitle    = {FAT*},
  pages        = {607--617},
  year         = {2020}
}

@article{laugel-corr17,
  author    = {Thibault Laugel and Marie{-}Jeanne Lesot and Christophe Marsala and
               Xavier Renard and Marcin Detyniecki},
  title     = {Inverse Classification for Comparison-based Interpretability in Machine Learning},
  journal   = {CoRR},
  volume    = {abs/1712.08443},
  year      = {2017}
}

@article{wachter-hjlt18,
  author    = {Sandra Wachter and Brent Mittelstadt and Chris Russell},
  title     = {Counterfactual Explanations Without Opening the Black Box:
               Automated Decisions and the {GDPR}},
  journal   = {Harvard Journal of Law \& Technology},
  volume    = {31},
  number    = {2},
  pages     = {841--887},
  year      = {2018}
}

@article{fischetti-cj18,
  author    = {Matteo Fischetti and
               Jason Jo},
  title     = {Deep neural networks and mixed integer linear optimization},
  journal   = {Constraints An Int. J.},
  volume    = {23},
  number    = {3},
  pages     = {296--309},
  year      = {2018},
  OPTurl       = {https://doi.org/10.1007/s10601-018-9285-6},
  OPTdoi       = {10.1007/s10601-018-9285-6},
  bibsource = {dblp computer science bibliography, https://dblp.org}
}

@article{imms-jsat19,
  author    = {Alexey Ignatiev and
               Antonio Morgado and
               Joao Marques{-}Silva},
  title     = {{RC2:} an Efficient {MaxSAT} Solver},
  journal   = {J. Satisf. Boolean Model. Comput.},
  volume    = {11},
  number    = {1},
  pages     = {53--64},
  year      = {2019},
}

@inproceedings{imms-sat18,
  author    = {Alexey Ignatiev and
               Antonio Morgado and
               Joao Marques{-}Silva},
  title     = {PySAT: {A} Python Toolkit for Prototyping with {SAT} Oracles},
  booktitle = {SAT},
  pages     = {428--437},
  year      = {2018},

}

@inproceedings{guestrin-kdd16a,
  author    = {Tianqi Chen and
               Carlos Guestrin},
  title     = {{XGBoost}: {A} Scalable Tree Boosting System},
  booktitle = {{KDD}},
  pages     = {785--794},
  year      = {2016}
}

@inproceedings{hcseyb-neurips16,
  author    = {Itay Hubara and
               Matthieu Courbariaux and
               Daniel Soudry and
               Ran El{-}Yaniv and
               Yoshua Bengio},
  title     = {Binarized Neural Networks},
  booktitle = {NIPS},
  pages     = {4107--4115},
  year      = {2016},
}

@book{molnar-bk20,
  title      = {Interpretable Machine Learning},
  author     = {Christoph Molnar},
  OPTnote       = {\url{https://christophm.github.io/interpretable-ml-book/}},
  note       = {\url{http://tiny.cc/6c76tz}},
  OPTyear       = {2019},
  year       = {2020},
  publisher  = {Leanpub},
  OPTsubtitle   = {A Guide for Making Black Box Models Explainable}
}

@article{friedman-tas01,
  author    = {Jerome H. Friedman},
  title     = {Greedy Function Approximation: A Gradient Boosting Machine},
  journal   = {The Annals of Statistics},
  number    = {5},
  pages     = {1189--1232},
  publisher = {Institute of Mathematical Statistics},
  volume    = {29},
  year      = {2001}
}

@inproceedings{iisms-aaai22,
  author    = {Alexey Ignatiev and
               Yacine Izza and
               Peter J. Stuckey and
               Joao Marques{-}Silva},
  title     = {Using {MaxSAT} for Efficient Explanations of Tree Ensembles},
  booktitle = {AAAI},
  pages     = {3776--3785},
  OPTpublisher = {{AAAI} Press},
  year      = {2022},
}

@article{rivest-ipl76,
  author    = {Laurent Hyafil and
               Ronald L. Rivest},
  title     = {Constructing Optimal Binary Decision Trees is {NP}-Complete},
  journal   = {Inf. Process. Lett.},
  volume    = {5},
  number    = {1},
  pages     = {15--17},
  year      = {1976},
}

@inproceedings{msi-aaai22,
  author       = {Joao Marques{-}Silva and
                  Alexey Ignatiev},
  title        = {Delivering Trustworthy {AI} through Formal {XAI}},
  booktitle    = {{AAAI}},
  pages        = {12342--12350},
  year         = {2022}
}

@inproceedings{lakkaraju-kdd16,
  author       = {Himabindu Lakkaraju and
                  Stephen H. Bach and
                  Jure Leskovec},
  title        = {Interpretable Decision Sets: {A} Joint Framework for Description and
                  Prediction},
  booktitle    = {{KDD}},
  pages        = {1675--1684},
  publisher    = {{ACM}},
  year         = {2016}
}

@inproceedings{hiicams-aaai22,
  author    = {Xuanxiang Huang and
               Yacine Izza and
               Alexey Ignatiev and
               Martin C. Cooper and
               Nicholas Asher and
               Joao Marques{-}Silva},
  title     = {Tractable Explanations for {d-DNNF} Classifiers},
  OPTbooktitle = {Thirty-Sixth {AAAI} Conference on Artificial Intelligence, {AAAI}
               2022, Thirty-Fourth Conference on Innovative Applications of Artificial
               Intelligence, {IAAI} 2022, The Twelveth Symposium on Educational Advances
               in Artificial Intelligence, {EAAI} 2022 Virtual Event, February 22
               - March 1, 2022},
  booktitle = {AAAI},
  pages     = {5719--5728},
  OPTpublisher = {{AAAI} Press},
  year      = {2022},
  OPTurl       = {https://ojs.aaai.org/index.php/AAAI/article/view/20514},
  bibsource = {dblp computer science bibliography, https://dblp.org}
}

@inproceedings{kwiatkowska-ijcai21,
  author    = {Emanuele La Malfa and
               Rhiannon Michelmore and
               Agnieszka M. Zbrzezny and
               Nicola Paoletti and
               Marta Kwiatkowska},
  OPTeditor    = {Zhi{-}Hua Zhou},
  title     = {On Guaranteed Optimal Robust Explanations for {NLP} Models},
  OPTbooktitle = {Proceedings of the Thirtieth International Joint Conference on Artificial
               Intelligence, {IJCAI} 2021, Virtual Event / Montreal, Canada, 19-27
               August 2021},
  booktitle = {IJCAI},
  pages     = {2658--2665},
  OPTpublisher = {ijcai.org},
  year      = {2021},
  OPTurl       = {https://doi.org/10.24963/ijcai.2021/366},
  OPTdoi       = {10.24963/ijcai.2021/366},
  bibsource = {dblp computer science bibliography, https://dblp.org}
}

@article{darwiche-jair21,
  author    = {Adnan Darwiche and Pierre Marquis},
  title     = {On Quantifying Literals in {Boolean} Logic and Its Applications to Explainable {AI}},
  journal   = {J. Artif. Intell. Res.},
  volume    = {72},
  pages     = {285--328},
  year      = {2021},
  OPTurl       = {},
  OPTdoi       = {},
  bibsource = {dblp computer science bibliography, https://dblp.org}
}

@inproceedings{rubin-aaai22,
  author    = {Niku Gorji and
               Sasha Rubin},
  title     = {Sufficient Reasons for Classifier Decisions in the Presence of Domain
               Constraints},
  booktitle = {{AAAI}},
  pages     = {5660--5667},
  year      = {2022}
}

@inproceedings{hinton-icml10,
  author       = {Vinod Nair and
                  Geoffrey Hinton},
  title        = {Rectified Linear Units Improve Restricted Boltzmann Machines},
  booktitle    = {{ICML}},
  pages        = {807--814},
  year         = {2010}
}

@article{lorini-logcom23,
  author       = {Xinghan Liu and
                  Emiliano Lorini},
  title        = {A unified logical framework for explanations in classifier systems},
  journal      = {J. Log. Comput.},
  volume       = {33},
  number       = {2},
  pages        = {485--515},
  year         = {2023}
}

@article{miller-corr24,
  author       = {Ronal Singh and
                  Tim Miller and
                  Liz Sonenberg and
                  Eduardo Velloso and
                  Frank Vetere and
                  Piers Howe and
                  Paul Dourish},
  title        = {An Actionability Assessment Tool for Explainable {AI}},
  journal      = {CoRR},
  volume       = {abs/2407.09516},
  year         = {2024}
}

@article{karimi-acmcs23,
  author       = {Amir{-}Hossein Karimi and
                  Gilles Barthe and
                  Bernhard Sch{\"{o}}lkopf and
                  Isabel Valera},
  title        = {A Survey of Algorithmic Recourse: Contrastive Explanations and Consequential
                  Recommendations},
  journal      = {{ACM} Comput. Surv.},
  volume       = {55},
  number       = {5},
  pages        = {95:1--95:29},
  year         = {2023}
}

@article{mccluskey-bstj56,
  title={Minimization of {Boolean} Functions},
  author={McCluskey, Edward J},
  journal={Bell system technical Journal},
  volume={35},
  number={6},
  pages={1417--1444},
  year={1956},
  publisher={Wiley Online Library}
}

@article{quine-amm55,
  title={A way to simplify truth functions},
  author={Quine, Willard V},
  journal={American mathematical monthly},
  pages={627--631},
  year={1955},
  publisher={JSTOR}
}

@article{quine-amm52,
  title={The problem of simplifying truth functions},
  author={Quine, Willard V},
  journal={American mathematical monthly},
  pages={521--531},
  year={1952},
  publisher={JSTOR}
}

@inproceedings{umans-focs98,
  author       = {Christopher Umans},
  title        = {The Minimum Equivalent {DNF} Problem and Shortest Implicants},
  OPTbooktitle    = {39th Annual Symposium on Foundations of Computer Science, {FOCS} '98,
                  November 8-11, 1998, Palo Alto, California, {USA}},
  booktitle    = {FOCS},
  pages        = {556--563},
  OPTpublisher    = {{IEEE} Computer Society},
  year         = {1998},
  OPTurl          = {https://doi.org/10.1109/SFCS.1998.743506},
  OPTdoi          = {10.1109/SFCS.1998.743506},
  bibsource    = {dblp computer science bibliography, https://dblp.org}
}

@article{umans-tcad06,
  author       = {Christopher Umans and
                  Tiziano Villa and
                  Alberto L. Sangiovanni{-}Vincentelli},
  title        = {Complexity of two-level logic minimization},
  journal      = {{IEEE} Trans. Comput. Aided Des. Integr. Circuits Syst.},
  volume       = {25},
  number       = {7},
  pages        = {1230--1246},
  year         = {2006},
  OPTurl          = {https://doi.org/10.1109/TCAD.2005.855944},
  OPTdoi          = {10.1109/TCAD.2005.855944},
  bibsource    = {dblp computer science bibliography, https://dblp.org}
}

@Misc{kelly-uci23,
  author       = {Markelle Kelly and
                  Rachel Longjohn and
                  Kolby Nottingham},
  title        = {The {UCI} Machine Learning Repository},
  howpublished = {\url{https://archive.ics.uci.edu}},
  year         = {2023},
  note         = {Accessed: 2026-01-14},
}

@article{romano-corr21,
  author       = {Joseph D. Romano and
                  Trang T. Le and
                  William G. La Cava and
                  John T. Gregg and
                  Daniel J. Goldberg and
                  Natasha L. Ray and
                  Praneel Chakraborty and
                  Daniel S. Himmelstein and
                  Weixuan Fu and
                  Jason H. Moore},
  title        = {{PMLB} v1.0: an open source dataset collection for benchmarking machine
                  learning methods},
  journal      = {CoRR},
  volume       = {abs/2012.00058v2},
  year         = {2021}
}

@article{olson-biodm17,
  author       = {Randal S. Olson and
                  William G. La Cava and
                  Patryk Orzechowski and
                  Ryan J. Urbanowicz and
                  Jason H. Moore},
  title        = {{PMLB:} a large benchmark suite for machine learning evaluation and
                  comparison},
  journal      = {BioData Min.},
  volume       = {10},
  number       = {1},
  pages        = {36:1--36:13},
  year         = {2017}
}

@incollection{stalnaker-theory68,
  author       = {Robert C. Stalnaker},
  title        = {A Theory of Conditionals},
  booktitle    = {Studies in Logical Theory},
  year         = {1968},
  pages        = {98--112}
}

@book{lewis-counterfactuals73,
  author       = {David Lewis},
  title        = {Counterfactuals},
  year         = {1973}
}

@inproceedings{sharygina-cav25,
  author       = {Faezeh Labbaf and
                  Tom{\'{a}}s Kol{\'{a}}rik and
                  Martin Blicha and
                  Grigory Fedyukovich and
                  Michael Wand and
                  Natasha Sharygina},
  title        = {Space Explanations of Neural Network Classification},
  booktitle    = {{CAV} {(3)}},
  pages        = {287--303},
  year         = {2025}
}

@article{craig-jsl57,
  author       = {William Craig},
  title        = {Three Uses of the {Herbrand-Gentzen} Theorem in Relating Model Theory
                  and Proof Theory},
  journal      = {J. Symb. Log.},
  volume       = {22},
  number       = {3},
  pages        = {269--285},
  year         = {1957}
}

@inproceedings{mcmillan-cav03,
  author       = {Kenneth L. McMillan},
  title        = {Interpolation and {SAT}-Based Model Checking},
  booktitle    = {{CAV}},
  pages        = {1--13},
  year         = {2003}
}

@inproceedings{darwiche-aaai22,
  author       = {Adnan Darwiche and
                  Chunxi Ji},
  title        = {On the Computation of Necessary and Sufficient Explanations},
  booktitle    = {{AAAI}},
  pages        = {5582--5591},
  year         = {2022}
}

@inproceedings{yu-aaai23,
  author       = {Jinqiang Yu and
                  Alexey Ignatiev and
                  Peter J. Stuckey and
                  Nina Narodytska and
                  Joao Marques{-}Silva},
  title        = {Eliminating the Impossible, Whatever Remains Must Be True: On Extracting
                  and Applying Background Knowledge in the Context of Formal Explanations},
  booktitle    = {{AAAI}},
  pages        = {4123--4131},
  year         = {2023}
}

@inproceedings{cooper-ecai23,
  author       = {Martin C. Cooper and
                  Leila Amgoud},
  title        = {Abductive Explanations of Classifiers Under Constraints: Complexity
                  and Properties},
  booktitle    = {{ECAI}},
  pages        = {469--476},
  year         = {2023}
}

@inproceedings{bassan-tacas23,
  author       = {Shahaf Bassan and
                  Guy Katz},
  title        = {Towards Formal {XAI:} Formally Approximate Minimal Explanations of
                  Neural Networks},
  booktitle    = {{TACAS} {(1)}},
  pages        = {187--207},
  year         = {2023}
}

@inproceedings{ji-neurips24,
  author       = {Xiayan Ji and
                  Anton Xue and
                  Eric Wong and
                  Oleg Sokolsky and
                  Insup Lee},
  title        = {{AR-Pro}: Counterfactual Explanations for Anomaly Repair with Formal
                  Properties},
  booktitle    = {NeurIPS},
  year         = {2024}
}

@inproceedings{amgoud-aamas26,
  author       = {Leila Amgoud and
                  Martin C. Cooper},
  title        = {Axiomatic Foundations of Counterfactual Explanations},
  booktitle    = {{AAMAS}},
  pages        = {282--290},
  year         = {2026}
}

@article{bassan-iclr26,
  author       = {Shahaf Bassan and
                  Xuanxiang Huang and
                  Guy Katz},
  title        = {Unifying Formal Explanations: {A} Complexity-Theoretic Perspective},
  journal      = {CoRR},
  volume       = {abs/2602.18160},
  year         = {2026}
}

@misc{gurobi,
  author = "Gurobi Optimization, LLC",
  title = "Gurobi Optimizer Reference Manual",
  year = 2026,
  url = "http://www.gurobi.com"
}

@article{artelt-corr19,
  author       = {Andr{\'{e}} Artelt and
                  Barbara Hammer},
  title        = {On the computation of counterfactual explanations - {A} survey},
  journal      = {CoRR},
  volume       = {abs/1911.07749},
  year         = {2019}
}

@article{guidotti-dmkd24,
  author       = {Riccardo Guidotti},
  title        = {Counterfactual explanations and how to find them: literature review
                  and benchmarking},
  journal      = {Data Min. Knowl. Discov.},
  volume       = {38},
  number       = {5},
  pages        = {2770--2824},
  year         = {2024}
}

@article{verma-acmcs24,
  author       = {Sahil Verma and
                  Varich Boonsanong and
                  Minh Hoang and
                  Keegan Hines and
                  John Dickerson and
                  Chirag Shah},
  title        = {Counterfactual Explanations and Algorithmic Recourses for Machine
                  Learning: {A} Review},
  journal      = {{ACM} Comput. Surv.},
  volume       = {56},
  number       = {12},
  pages        = {312:1--312:42},
  year         = {2024}
}

@article{guidotti-corr18,
  author       = {Riccardo Guidotti and
                  Anna Monreale and
                  Salvatore Ruggieri and
                  Dino Pedreschi and
                  Franco Turini and
                  Fosca Giannotti},
  title        = {Local Rule-Based Explanations of Black Box Decision Systems},
  journal      = {CoRR},
  volume       = {abs/1805.10820},
  year         = {2018}
}

@inproceedings{upadhyay-neurips21,
  author       = {Sohini Upadhyay and
                  Shalmali Joshi and
                  Himabindu Lakkaraju},
  title        = {Towards Robust and Reliable Algorithmic Recourse},
  booktitle    = {NeurIPS},
  pages        = {16926--16937},
  year         = {2021}
}

\clearpage
\appendix

\section{Experimental Details and Complete Results}\label{sec:app-experiments}

This appendix gives the details of the experimental results omitted
from \autoref{sec:res}.
%
% The supplementary material additionally contains the datasets, trained
% models, logs, raw CSV files, and scripts used to produce the results.

\subsection{Benchmarks and Models}\label{sec:app-benchmarks}

All datasets are public tabular classification benchmarks obtained
from UCI or PMLB~\citep{kelly-uci23,olson-biodm17,romano-corr21}.
\autoref{tab:benchmarks} provides the details of the datasets and the
XGBoost and ReLU models trained in both experiments.
Every original input column is treated as numerical.
The same fixed 80\%/20\% split has been applied when training both
models.

For the XGBoost experiment, every model uses 50 trees per class, an
80/20 train/test split, and the maximum depth shown in
\autoref{tab:benchmarks}.
The split uses the default seed value 7.
The reported propositional encodings have 4--10183 variables and
304--102611 clauses.
For the neural experiment, every feature is scaled to $[0,1]$.
A fully connected \texttt{sklearn} \texttt{MLPClassifier} with ReLU
hidden layers $(16,8)$ is trained on the training instances only using
Adam, learning rate 0.005, maximum 1000 iterations, and seed 1234.
Note that both XGBoost and ReLU models are trained on exactly the same
training/test data, to illustrate the performance of CFXp on both.

\begin{table*}[!ht]
\centering
% \scriptsize
\caption{Benchmark and trained-model details. $n$, $|\varphi|$, and
$|\mbb{K}|$ are the numbers of rows, features, and classes. $d$, tr.,
and te. denote XGBoost's maximum tree depth and train/test accuracy;
$p$ is the ReLU model's number of trainable parameters.}
\label{tab:benchmarks}
\begin{adjustbox}{center}
\begin{tabular}{lrrrrrrrrr}
\toprule
Dataset & $n$ & $|\varphi|$ & $|\mbb{K}|$ & $d$ & XGB tr. & XGB te. & $p$ & ReLU tr. & ReLU te. \\
\midrule
% Automatically generated from benchmark, XGBoost result, and ReLU model artifacts.
ann-thyroid      &  7200 & 21 &  3 & 3 & 1.000 & 1.000 &  515 & 0.989 & 0.981 \\
appendicitis     &   106 &  7 &  2 & 3 & 0.990 & 0.910 &  273 & 0.905 & 0.864 \\
biodegradation   &  1055 & 41 &  2 & 3 & 0.970 & 0.870 &  817 & 0.922 & 0.872 \\
divorce          &   170 & 54 &  2 & 3 & 1.000 & 1.000 & 1025 & 1.000 & 1.000 \\
ecoli            &   327 &  7 &  5 & 3 & 1.000 & 0.850 &  309 & 0.904 & 0.909 \\
glass2           &   163 &  9 &  2 & 4 & 1.000 & 0.880 &  305 & 0.892 & 0.848 \\
ionosphere       &   351 & 34 &  2 & 3 & 1.000 & 0.930 &  705 & 0.989 & 0.944 \\
pendigits        & 10992 & 16 & 10 & 3 & 1.000 & 0.990 &  498 & 0.992 & 0.984 \\
promoters        &   106 & 58 &  2 & 3 & 1.000 & 1.000 & 1089 & 1.000 & 0.909 \\
segmentation     &   210 & 19 &  7 & 3 & 1.000 & 0.950 &  519 & 0.994 & 0.857 \\
shuttle          & 58000 &  9 &  7 & 3 & 1.000 & 1.000 &  359 & 0.998 & 0.997 \\
sonar            &   208 & 60 &  2 & 3 & 1.000 & 0.860 & 1121 & 1.000 & 0.833 \\
spambase         &  4601 & 57 &  2 & 4 & 0.980 & 0.960 & 1073 & 0.990 & 0.939 \\
texture          &  5500 & 40 & 11 & 3 & 1.000 & 0.980 &  891 & 0.992 & 0.985 \\
threeOf9         &   512 &  9 &  2 & 3 & 1.000 & 1.000 &  305 & 1.000 & 1.000 \\
twonorm          &  7400 & 20 &  2 & 3 & 1.000 & 0.970 &  481 & 0.979 & 0.977 \\
vowel            &   990 & 13 & 11 & 4 & 1.000 & 0.930 &  459 & 0.973 & 0.894 \\
wdbc             &   569 & 30 &  2 & 4 & 1.000 & 0.970 &  641 & 0.982 & 0.991 \\
wine-recognition &   178 & 13 &  3 & 3 & 1.000 & 0.970 &  387 & 1.000 & 0.972 \\
wpbc             &   194 & 33 &  2 & 4 & 1.000 & 0.740 &  689 & 1.000 & 0.692 \\
zoo              &   101 & 16 &  7 & 4 & 1.000 & 0.830 &  471 & 1.000 & 1.000 \\
\\[-1.5ex]
\bottomrule
\end{tabular}
\end{adjustbox}
\end{table*}

\subsection{Tree Ensembles: CWC Reference Experiment}\label{sec:app-trees}

The instances are sampled once by the original experiment with
\texttt{random.seed(1234)}: up to 200 instances per dataset (or every
instance for smaller datasets).
This gives 3755 queries.
AXp, CXp, and CFXp extraction uses the same MaxSAT encoding and
incremental RC2 solver~\cite{imms-sat18,imms-jsat19} as the earlier
MaxSAT study~\cite{iisms-aaai22}.
A CFXp run first extracts a CXp and witness, then extracts an AXp for
this witness, as described in \autoref{sec:comp}.
The five-minute limit applies independently to each formal explanation
run; none of the AXp, CXp, or CFXp runs timed out.
Reported experiment times are process CPU times and exclude model
encoding that is done once per dataset.

CWC builds on exhaustive enumeration of closest worlds rather than a
single-witness algorithm.
For each query, it enumerates all minimum-cardinality CXps and all
their admissible witnesses over the finite partition induced by tree
thresholds.
Of the 3755 CWC computations, 3327 completed and 428 reached the
300-second limit.
Its completed runs enumerate 1--17697 worlds (mean/median 117.770/19).
The CWC mean in the main paper, 51.872 seconds, uses 300 seconds for
every unfinished run.
It is consequently a lower bound on the uncapped mean.

\begin{table}[t]
\centering
% \small
\caption{Aggregate formal explanation extraction results, for both
XGBoost and ReLU models. Size and time report mean/median.
CWC was run only on XGBoost models as it cannot be used for
infinite feature domains. CWC size is the number of enumerated worlds on its 3327
completed runs; its time includes 300-second caps for 428 unfinished runs.}
\label{tab:formal-aggregates}
\begin{tabular}{llrr}
\toprule
Model & Explanation & Size & Time (s) \\
\midrule
XGBoost & AXp & 9.043/{\color{white}0}8 & 1.529/0.266 \\
XGBoost & CXp ($|\gamma|$) & 1.972/{\color{white}0}1 & 0.783/0.184 \\
XGBoost & CFXp support ($|\delta|$) & 8.240/{\color{white}0}7 & -- \\
XGBoost & CFXp certificate ($|\gamma\cup\delta|$) & 10.212/{\color{white}0}8 & 2.098/0.500 \\
XGBoost & CWC worlds & 117.770/19 & 51.872/2.168 \\
\midrule
ReLU & AXp & 19.734/15 & 0.238/0.075 \\
ReLU & CXp ($|\gamma|$) & 2.064/{\color{white}0}1 & 0.064/0.033 \\
ReLU & CFXp support ($|\delta|$) & 20.577/17 & -- \\
ReLU & CFXp certificate ($|\gamma\cup\delta|$) & 22.641/19 & 0.310/0.076 \\
\bottomrule
\end{tabular}
\end{table}

\begin{figure*}[t]
\centering
\begin{subfigure}{.49\textwidth}
  \includegraphics[width=\linewidth]{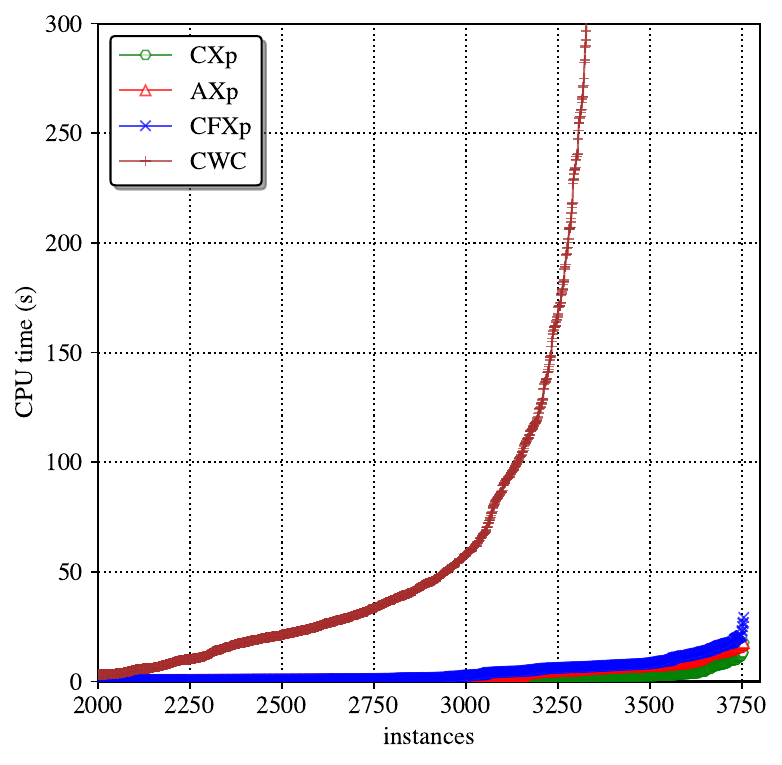}
  \caption{Linear scale.}
\end{subfigure}\hfill
\begin{subfigure}{.49\textwidth}
  \includegraphics[width=\linewidth]{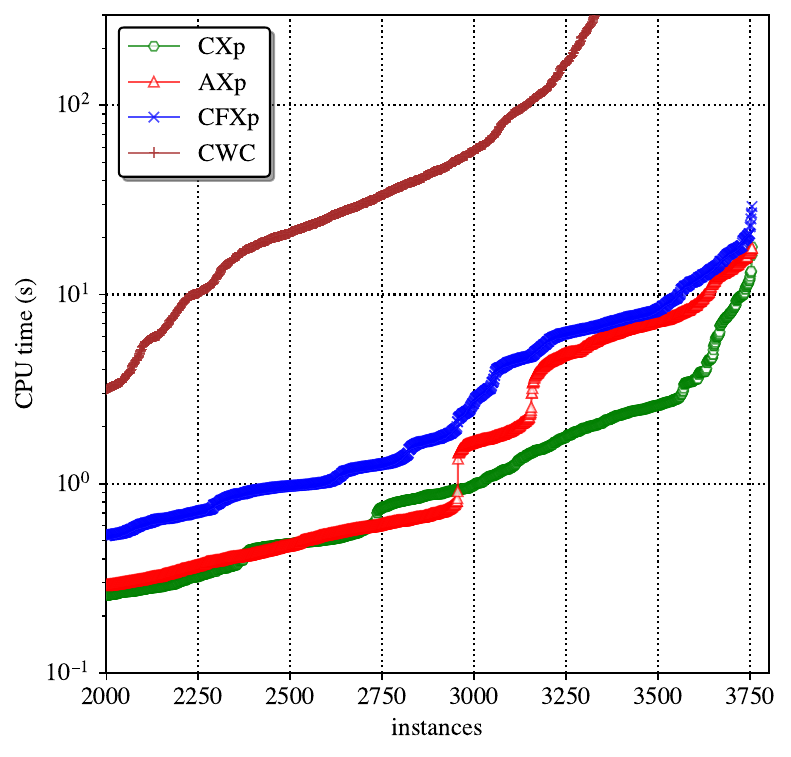}
  \caption{Log scale.}
\end{subfigure}
\caption{Cactus plots for tree-ensemble extraction.  CWC includes capped
300-second timeout observations.}
\label{fig:tree-cactus}
\end{figure*}

\begin{figure*}[t]
\centering
\begin{subfigure}{.32\textwidth}
  \includegraphics[width=\linewidth]{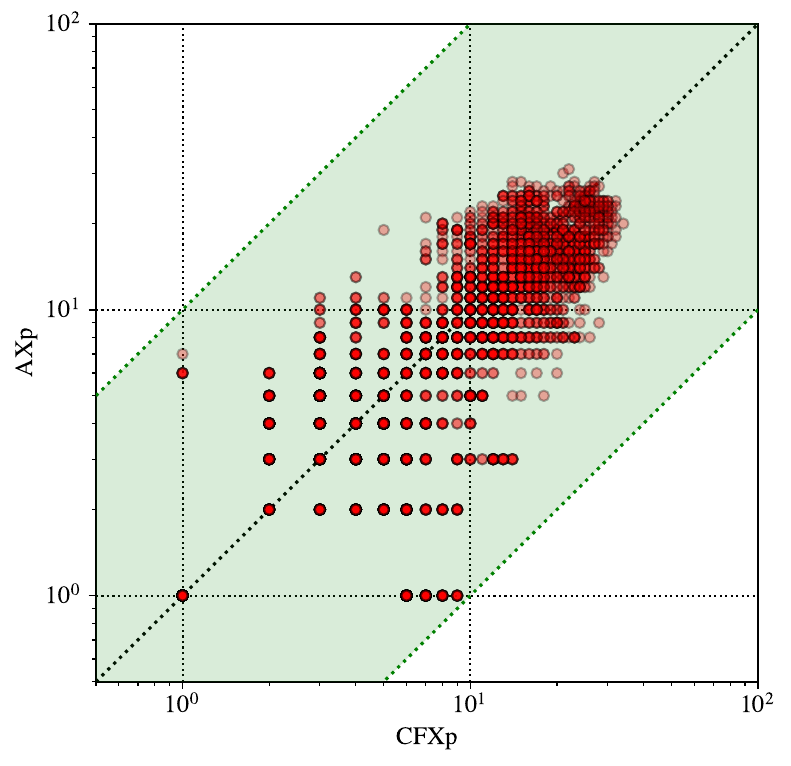}
  \caption{AXp.}
\end{subfigure}\hfill
\begin{subfigure}{.32\textwidth}
  \includegraphics[width=\linewidth]{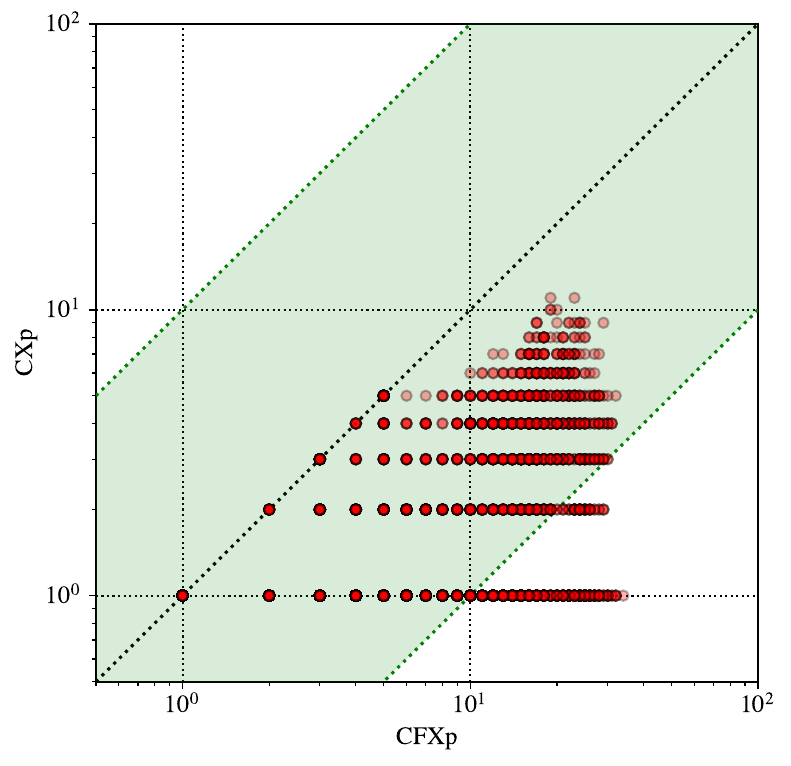}
  \caption{CXp.}
\end{subfigure}\hfill
\begin{subfigure}{.32\textwidth}
  \includegraphics[width=\linewidth]{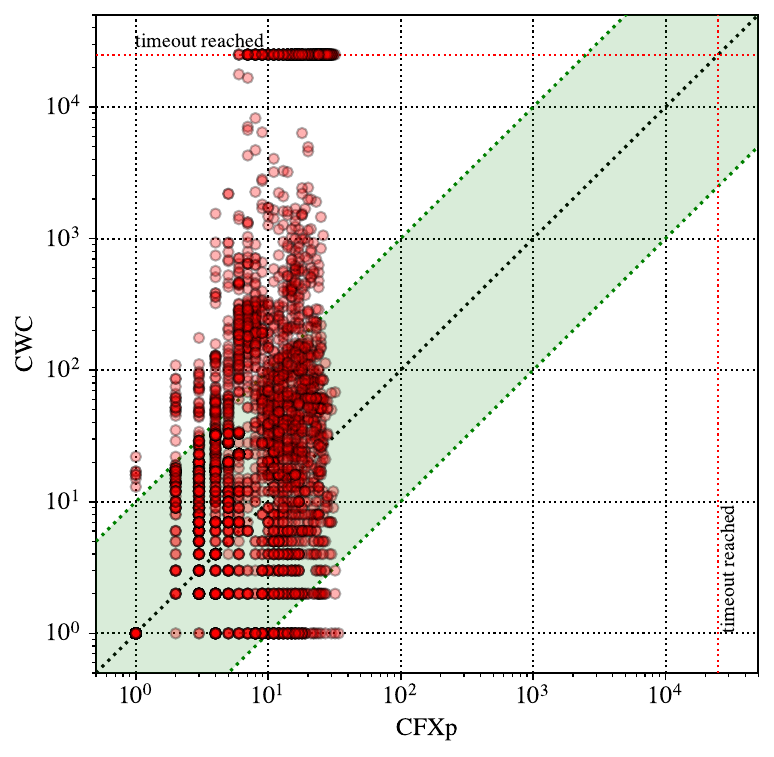}
  \caption{CWC.}
\end{subfigure}
\\[1ex]
\begin{subfigure}{.32\textwidth}
  \includegraphics[width=\linewidth]{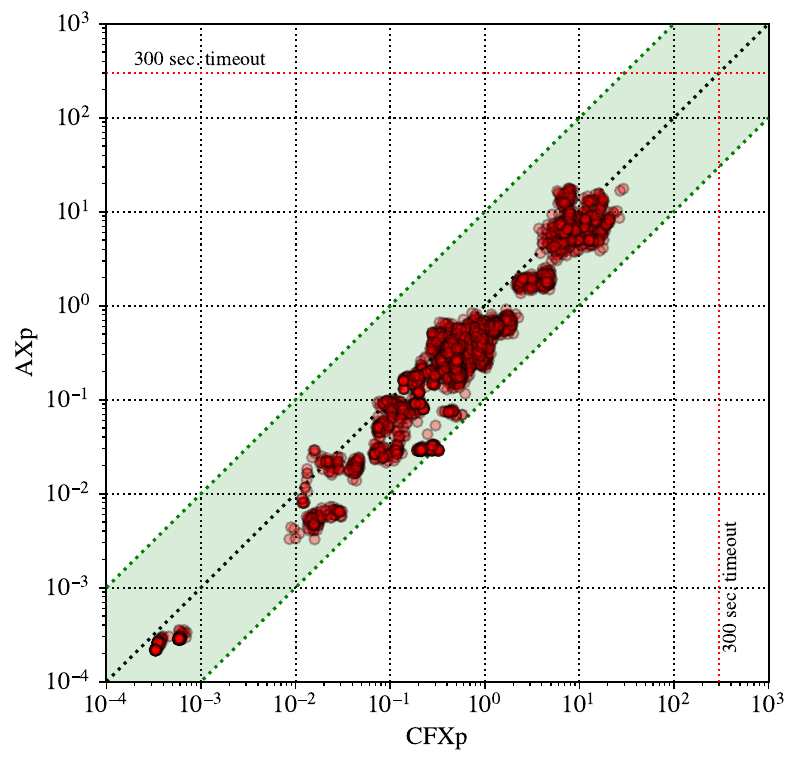}
  \caption{AXp.}
\end{subfigure}\hfill
\begin{subfigure}{.32\textwidth}
  \includegraphics[width=\linewidth]{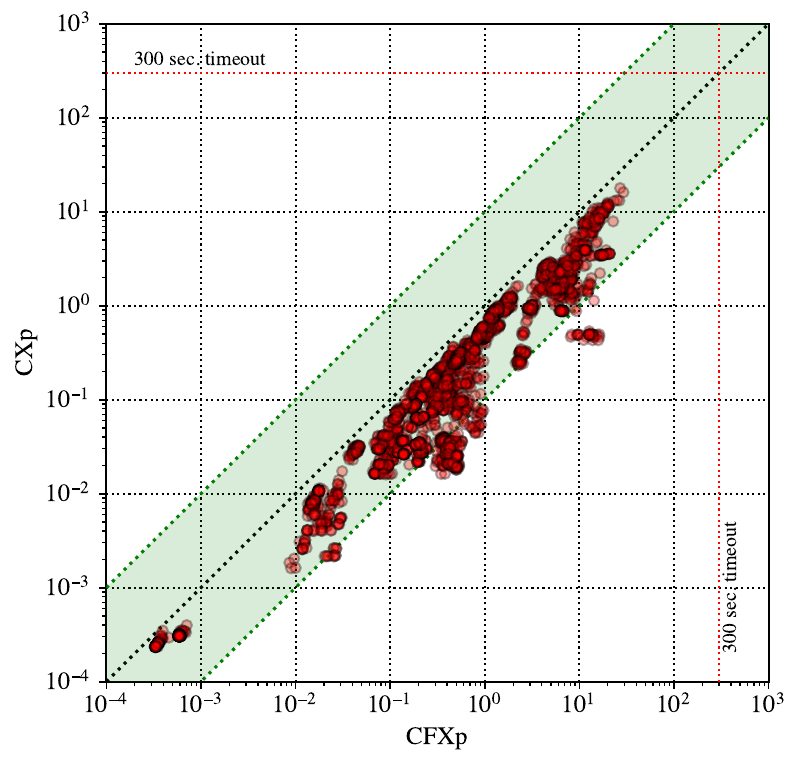}
  \caption{CXp.}
\end{subfigure}\hfill
\begin{subfigure}{.32\textwidth}
  \includegraphics[width=\linewidth]{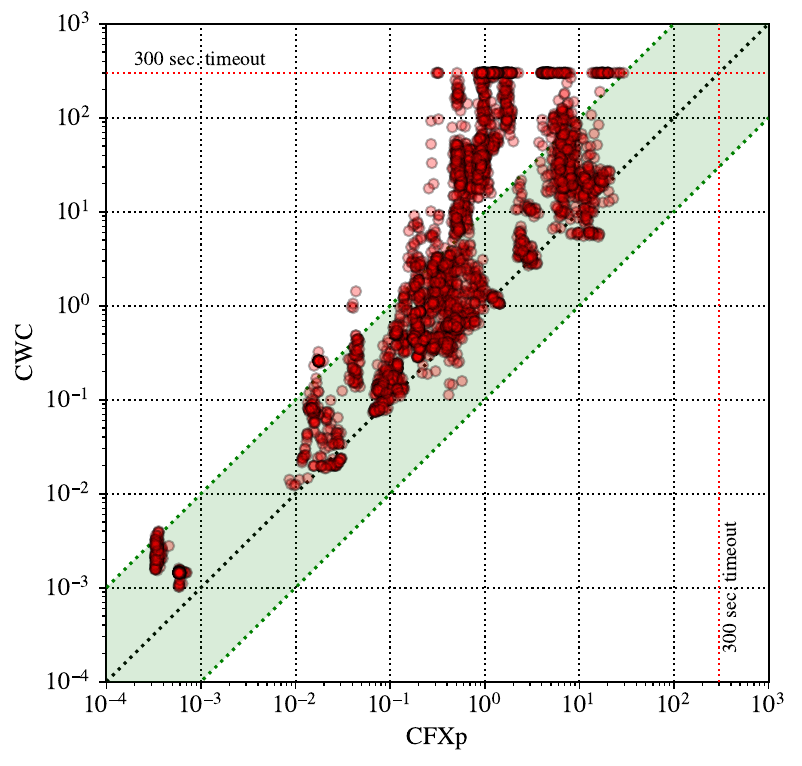}
  \caption{CWC.}
\end{subfigure}
\caption{Per-instance XGBoost comparisons. Top: explanation sizes,
with CFXp measured by the joint certificate $|\gamma\cup\delta|$;
bottom: CPU times.}
\label{fig:tree-scatter}
\end{figure*}

\autoref{fig:tree-cactus} makes the cost of CWC visible.
Although many CWC queries finish quickly, a larger number reaches the
time limit, making CWC fall behind all the other formal approaches.
The paired plots in \autoref{fig:tree-scatter} give the full
explanation size and runtime details behind the aggregate values shown
in \autoref{sec:cwc-res}.
They should not be read as a comparison of interchangeable output
objects: CWC represents all closest worlds, whereas CFXp returns one
witness together with a regional certificate.

\subsection{Point-Explainer Setup for Tree Ensembles}\label{sec:app-tree-points}

The point methods use precisely the same trained XGBoost models and
3755 queries.
DiCE 0.12 uses its random method with 1000 samples, post-hoc linear
sparsity enhancement 0.1, and deterministic seed
$1234+100\cdot\mathit{instance\_id}+\mathit{target\_class}$.
For each multiclass query, DiCE is run for every alternative class
such that the model-verified point with lexicographically minimum
$L_0$, normalized $L_1$, and normalized $L_2$ distance is picked.
Its time includes all target searches.

Growing Spheres uses the original authors' implementation at commit
\href{https://github.com/thibaultlaugel/growingspheres/commit/894ce969f39f4f927f1732e569fb5e7ac6f770a9}{\texttt{894ce96}},
with its documented defaults: 2000 samples per layer, ring layers,
radius 0.1, decrease factor 10, untargeted search, and sparse
projection.

Wachter uses
\href{https://rasbt.github.io/mlxtend/user_guide/evaluate/create_counterfactual/}{\texttt{mlxtend.evaluate.create\_counterfactual}}
from \texttt{mlxtend}~0.24.0.
It scales each feature to $[0,1]$, tries every alternative target and
$\lambda\in\{0.1,1,10,100,1000\}$, and asks for target probability 1.
The implementation's random initial point is restricted to rows
already predicted as the intended target.
Note that Wachter optimizes normalized $L_1$, not feature sparsity.

\begin{figure*}[t]
\centering
\begin{subfigure}{.49\textwidth}
  \includegraphics[width=\linewidth]{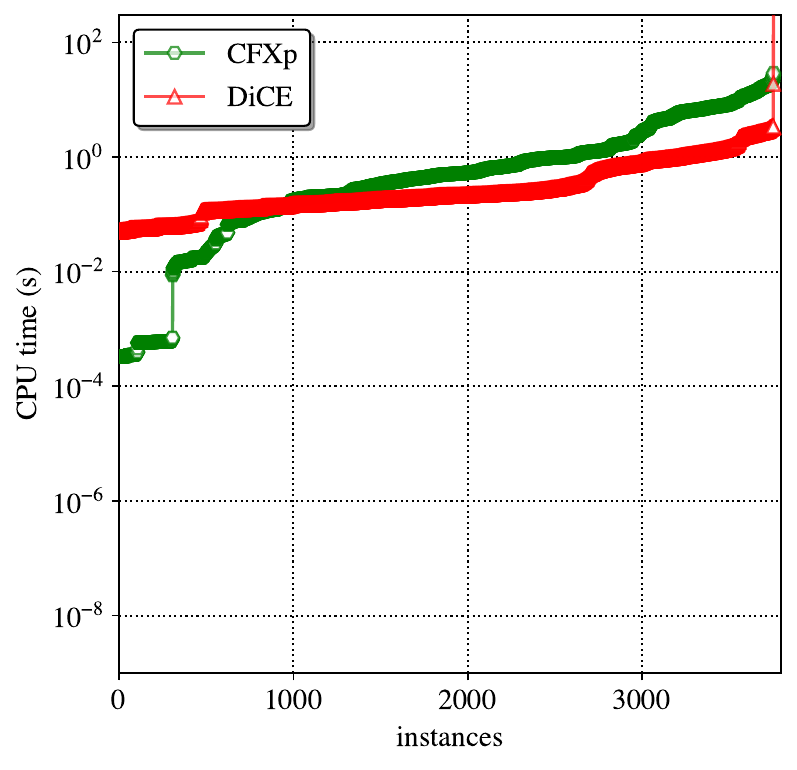}
  \caption{All datasets: CFXp and DiCE.}
\end{subfigure}\hfill
\begin{subfigure}{.49\textwidth}
  \includegraphics[width=\linewidth]{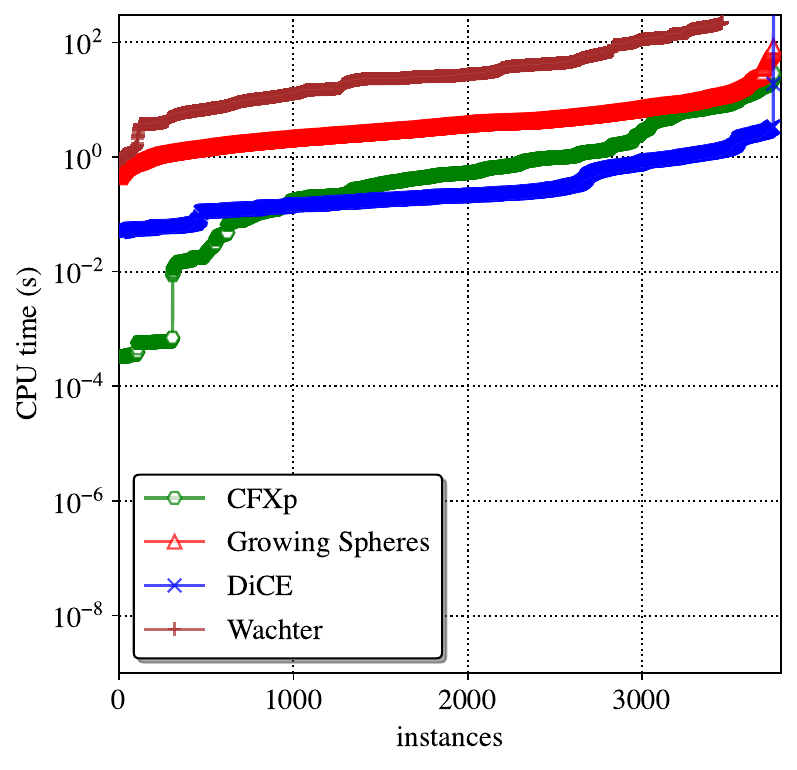}
  \caption{All datasets: all point methods.}
\end{subfigure}
\caption{XGBoost point-comparison runtime cactus plots. A point method that
does not return a model-verified point is placed at the 300-second plot
boundary.}
\label{fig:xgb-point-times}
\end{figure*}

\begin{figure*}[t]
\centering
\begin{subfigure}{.32\textwidth}
  \includegraphics[width=\linewidth]{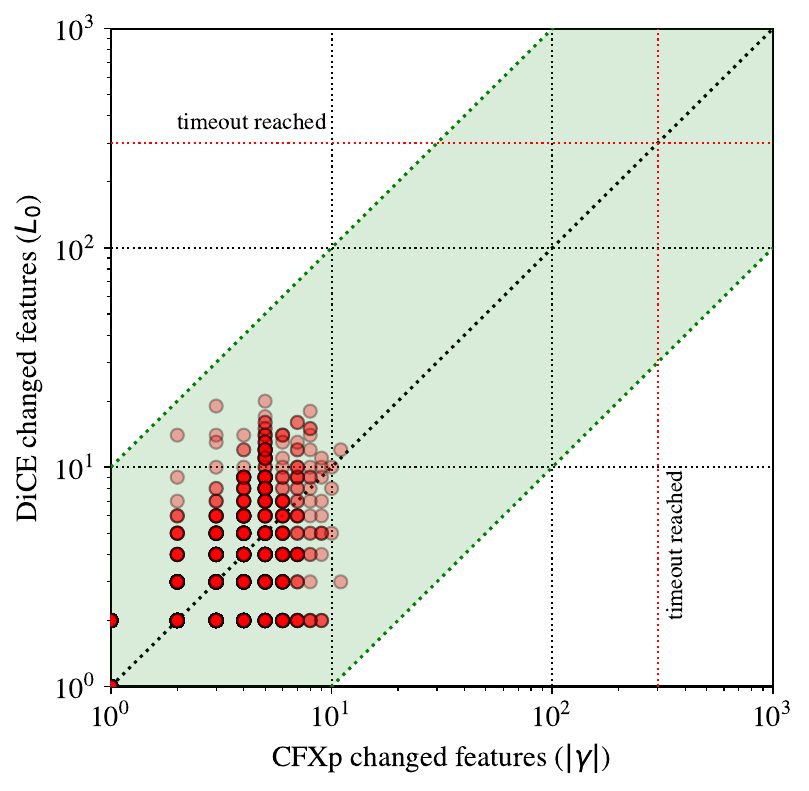}
  \caption{DiCE.}
\end{subfigure}\hfill
\begin{subfigure}{.32\textwidth}
  \includegraphics[width=\linewidth]{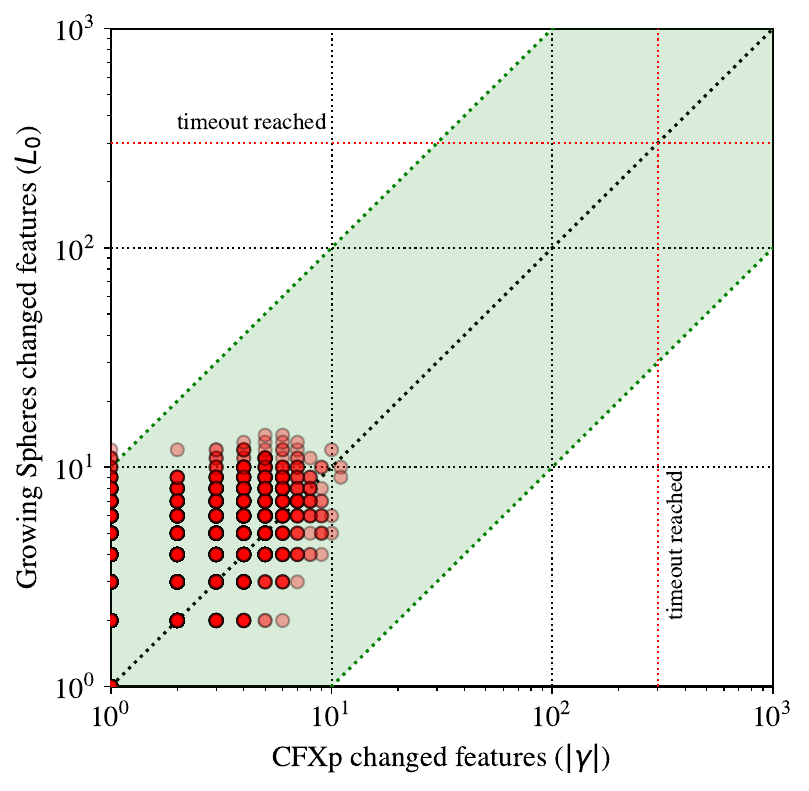}
  \caption{Growing Spheres.}
\end{subfigure}\hfill
\begin{subfigure}{.32\textwidth}
  \includegraphics[width=\linewidth]{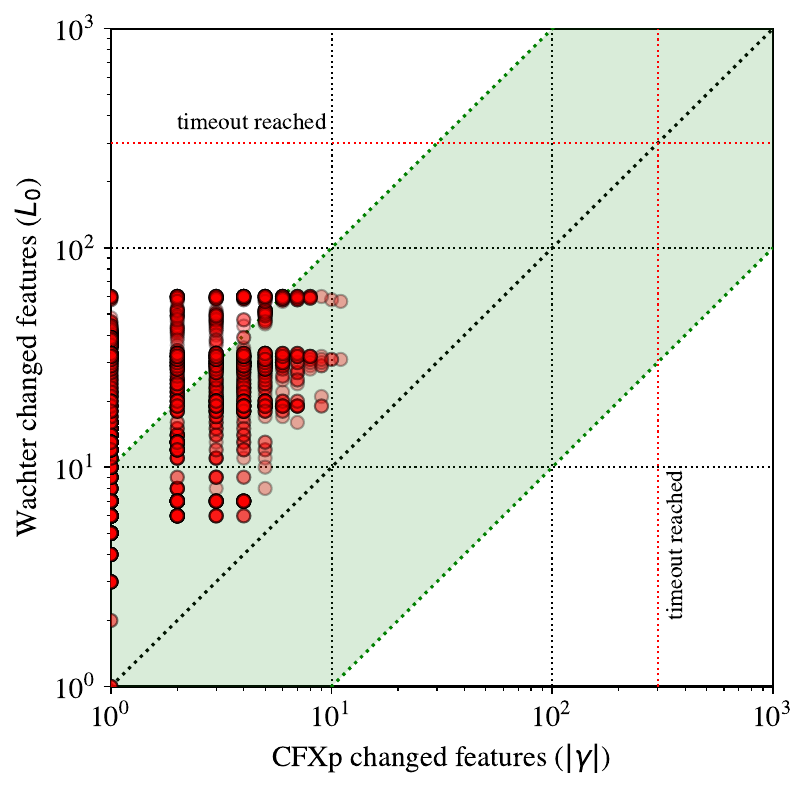}
  \caption{Wachter.}
\end{subfigure}
\caption{XGBoost paired changed-feature statistics. Only model-verified point
outputs are plotted; the horizontal axis is CFXp changed-feature set $|\gamma|$.}
\label{fig:xgb-point-sizes}
\end{figure*}

Every input column is numerical and all three point methods are run on
all 3755 queries.
Every returned point is re-evaluated by XGBoost and must be finite and
within the observed feature ranges.
Only these model-verified points contribute to the paired size
statistics shown in the paper.
CPU time is plotted in \autoref{fig:xgb-point-times}.
Initialization is excluded from both baseline and MaxSAT runtimes.
Explanation size comparison is detailed in the scatter plots of
\autoref{fig:xgb-point-sizes}.

\subsection{ReLU Networks Experiment}\label{sec:app-relu}

The neural experiment explains the decisions made by the trained ReLU
models shown in \autoref{tab:benchmarks} on the same raw query instances.
Each original feature corresponds to one numerical MILP input with
bounds $[0,1]$.
The backend follows the Fischetti--Jo ReLU formulation in
mixed-integer linear programming (MILP)~\cite{fischetti-cj18}.
For a ReLU, the encoding introduces a binary activation indicator and
positive/negative parts.
Class change is tested against each alternative output score with
margin $10^{-4}$.
Gurobi~\cite{gurobi} uses one thread, a 60-second per-oracle-call
limit, feasibility and integer-feasibility tolerances $10^{-8}$, and
disabled dual reductions.
All 3755 formal explanations are successfully extracted, and every
witness returned along the way for the SAT oracle calls is
re-evaluated with the sklearn model.

Neural CXp, AXp, and CFXp extraction takes mean/median 0.064/0.033,
0.238/0.075, and 0.310/0.076 seconds, respectively.
The corresponding mean/median oracle-call counts are 27.979/21 for
CXps and 24.916/18 for AXp supports.
The CFXp change set $\gamma$ has range 1--25 and mean/median 2.064/1;
the support $\delta$ and complete certificate $\gamma\cup\delta$ have
ranges 0--59 and 1--60 and mean/median sizes 20.577/17 and 22.641/19.
The larger supports are consistent with ordinary neural AXps, whose
size is 19.734/15 on average/median.

\begin{figure*}[t]
\centering
\begin{subfigure}{.31\textwidth}
  \includegraphics[width=\linewidth]{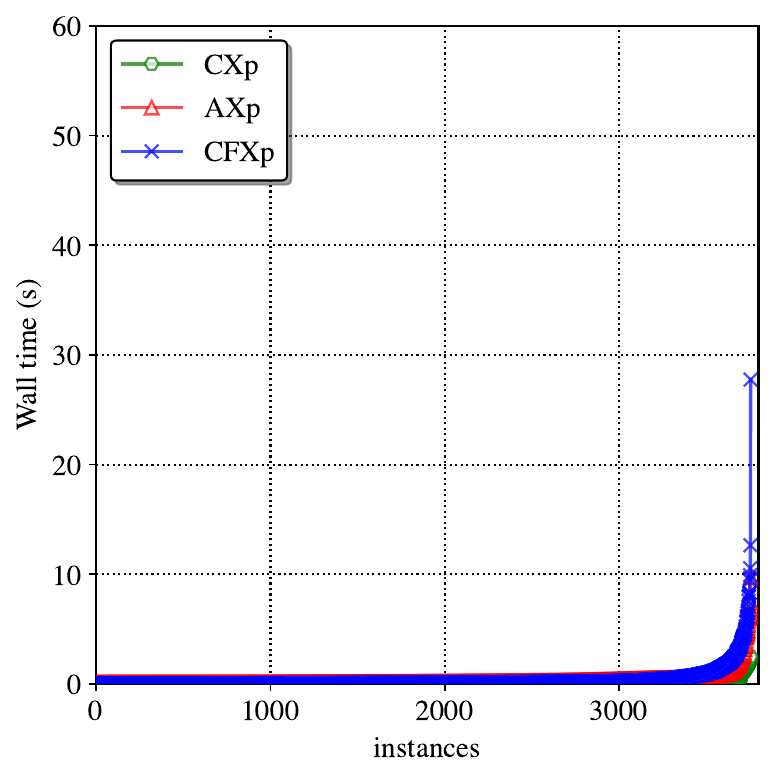}
  \caption{Linear cactus plot.}
\end{subfigure}\hfill
\begin{subfigure}{.32\textwidth}
  \includegraphics[width=\linewidth]{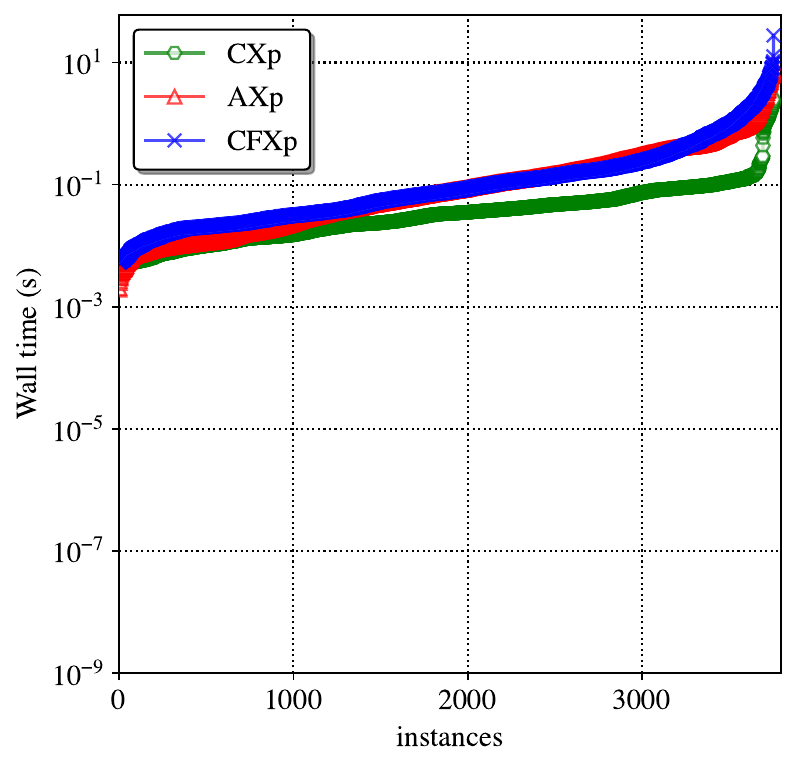}
  \caption{Log cactus plot.}
\end{subfigure}\hfill
\begin{subfigure}{.32\textwidth}
  \includegraphics[width=\linewidth]{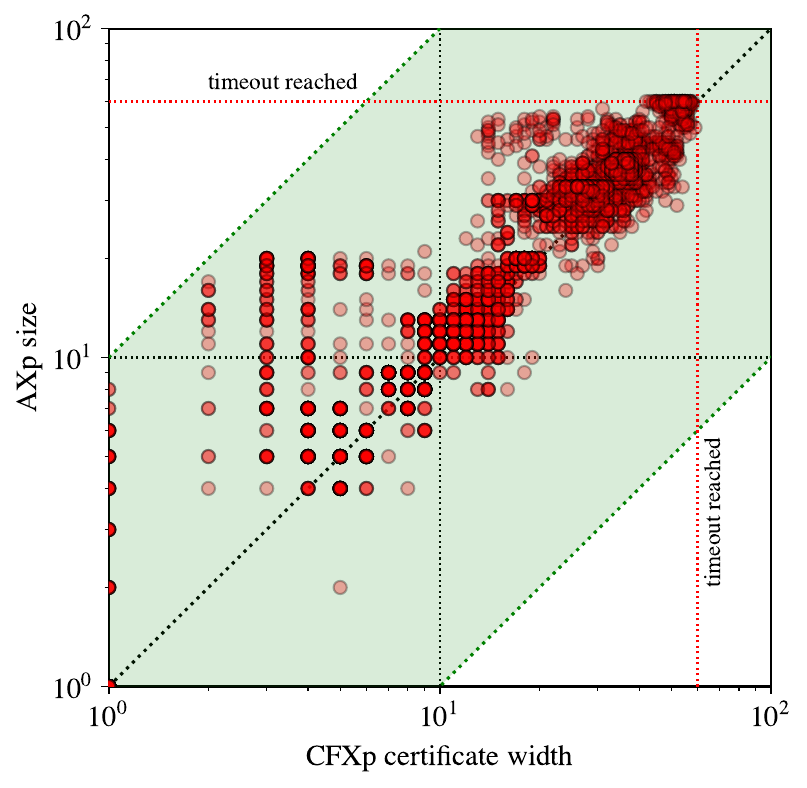}
  \caption{CFXp $|\gamma\cup\delta|$ vs.\ AXp size.}
\end{subfigure}
\\[1ex]
\begin{subfigure}{.32\textwidth}
  \includegraphics[width=\linewidth]{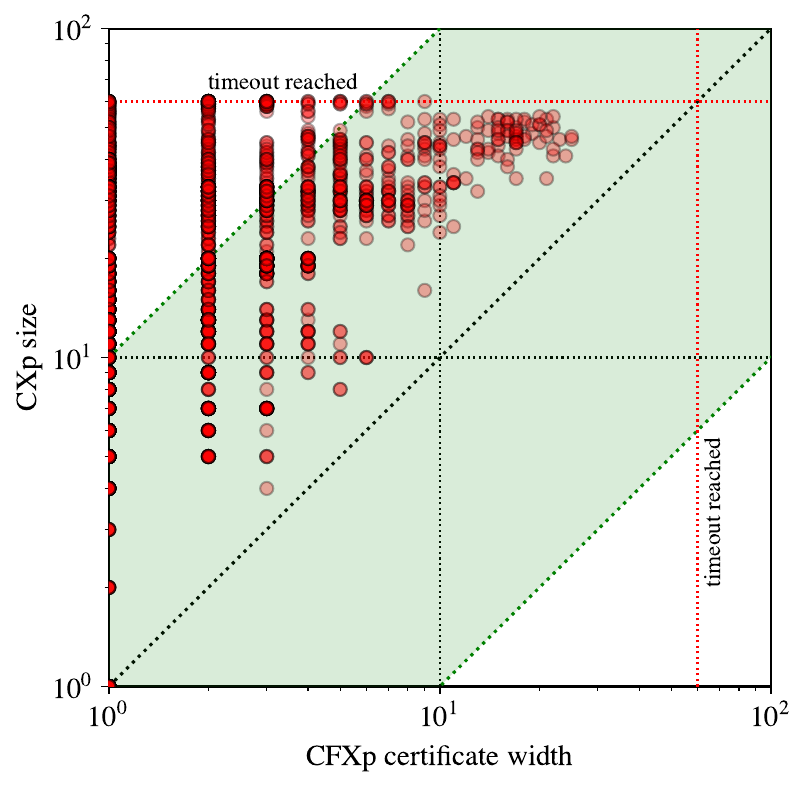}
  \caption{CFXp $|\gamma\cup\delta|$ vs.\ CXp size.}
\end{subfigure}\hfill
\begin{subfigure}{.32\textwidth}
  \includegraphics[width=\linewidth]{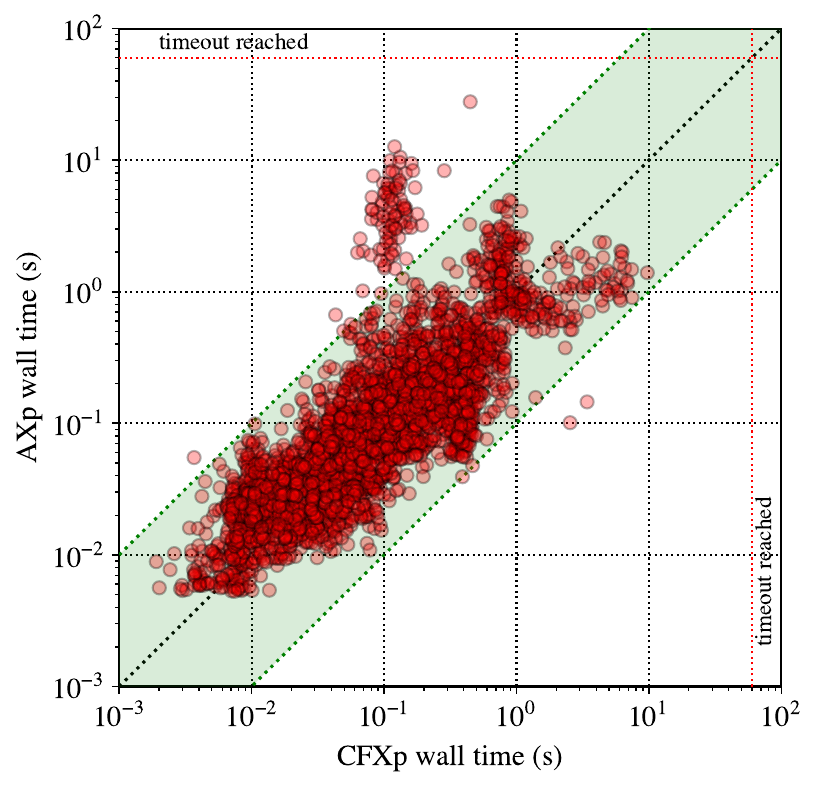}
  \caption{CFXp versus AXp time.}
\end{subfigure}\hfill
\begin{subfigure}{.32\textwidth}
  \includegraphics[width=\linewidth]{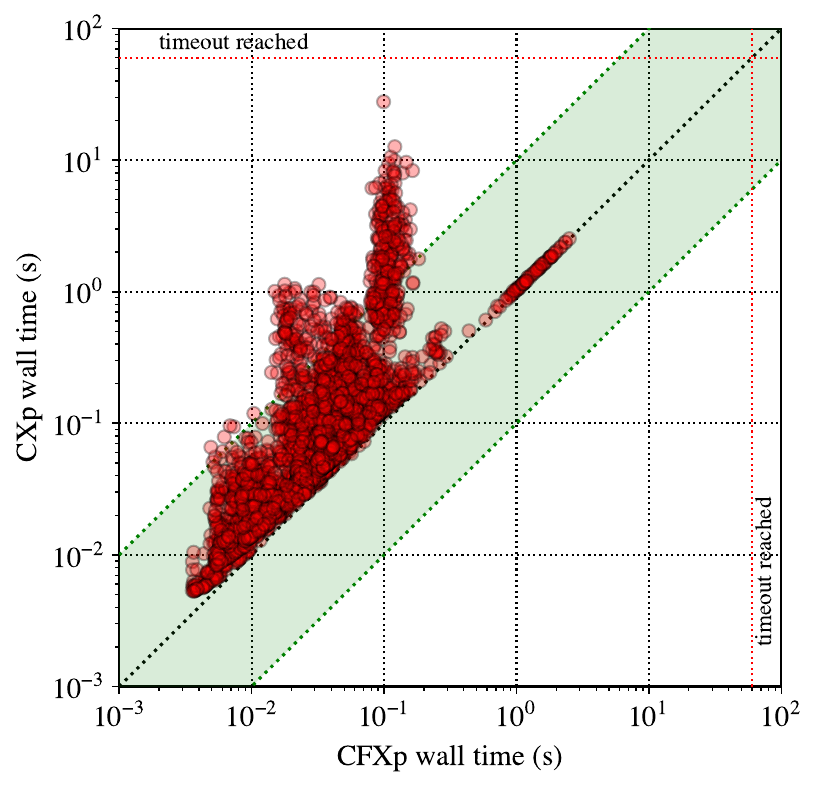}
  \caption{CFXp versus CXp time.}
\end{subfigure}
\caption{Complete ReLU formal-extraction results, in the same style as
the tree-ensemble plots. Times are single-threaded wall-clock times.}
\label{fig:relu-distributions}
\end{figure*}

As for the baselines, DiCE uses the same configuration as above and
selects the best verified alternative target.
Growing Spheres again uses the pinned authors' implementation and
Wachter again uses \texttt{mlxtend} 0.24.0 with the same target and
$\lambda$ sweep.
All three methods are run on the same 3755 queries.
Candidate points are checked against the sklearn classifier, feature
ranges, and the requested target.
Unlike the XGBoost experiment, the times in \autoref{tab:point-res}
are wall-clock times for both neural CFXp extraction and point
baselines.
The full per-instance raw results and per-dataset summaries are
provided.

\begin{figure*}[t]
\centering
\begin{subfigure}{.49\textwidth}
  \includegraphics[width=\linewidth]{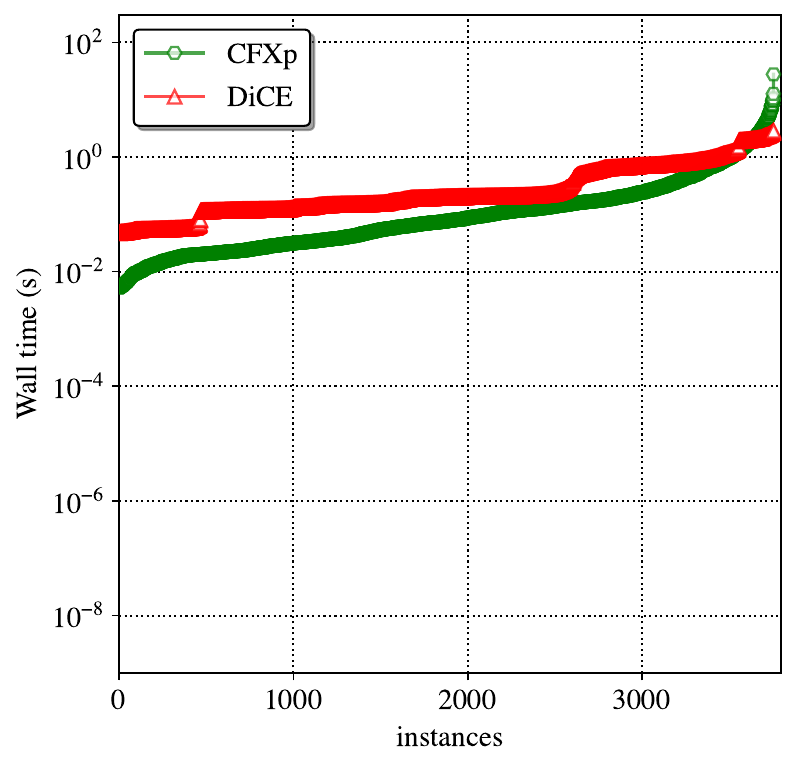}
  \caption{All datasets: CFXp and DiCE.}
\end{subfigure}\hfill
\begin{subfigure}{.49\textwidth}
  \includegraphics[width=\linewidth]{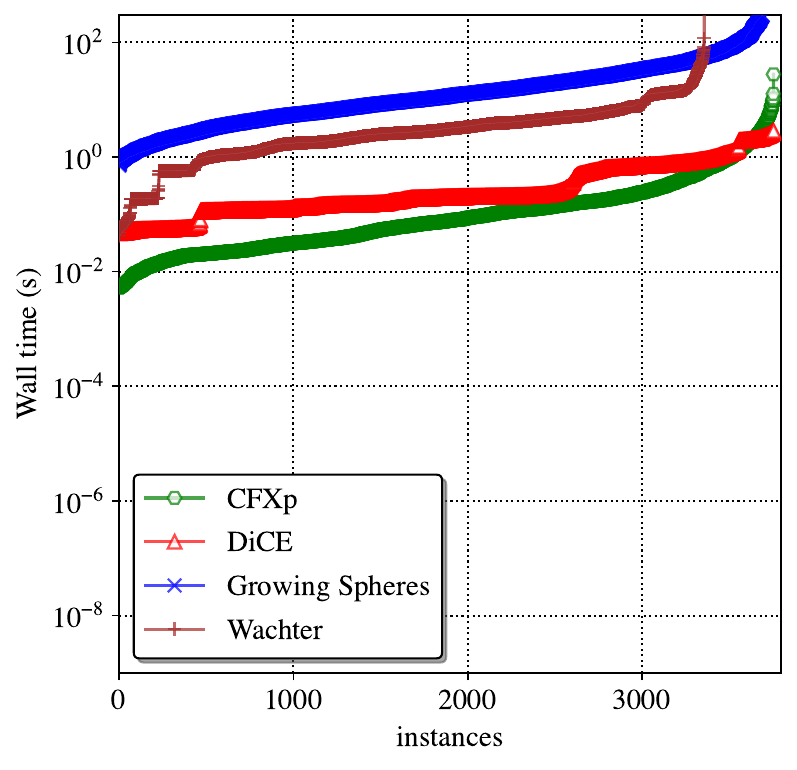}
  \caption{All datasets: all point methods.}
\end{subfigure}
\caption{Runtime cactus plots for ReLU benchmarks. Calls producing points
  that are not verified by the model are placed at the 300-second plot
  boundary.}
\label{fig:relu-point-times}
\end{figure*}

\begin{figure*}[t]
\centering
\begin{subfigure}{.32\textwidth}
  \includegraphics[width=\linewidth]{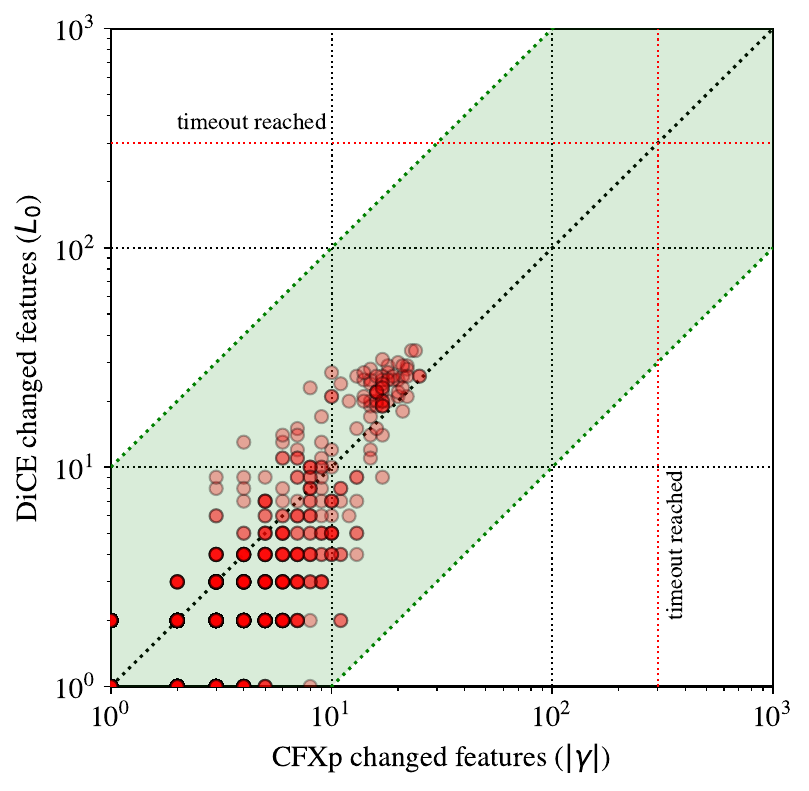}
  \caption{DiCE.}
\end{subfigure}\hfill
\begin{subfigure}{.32\textwidth}
  \includegraphics[width=\linewidth]{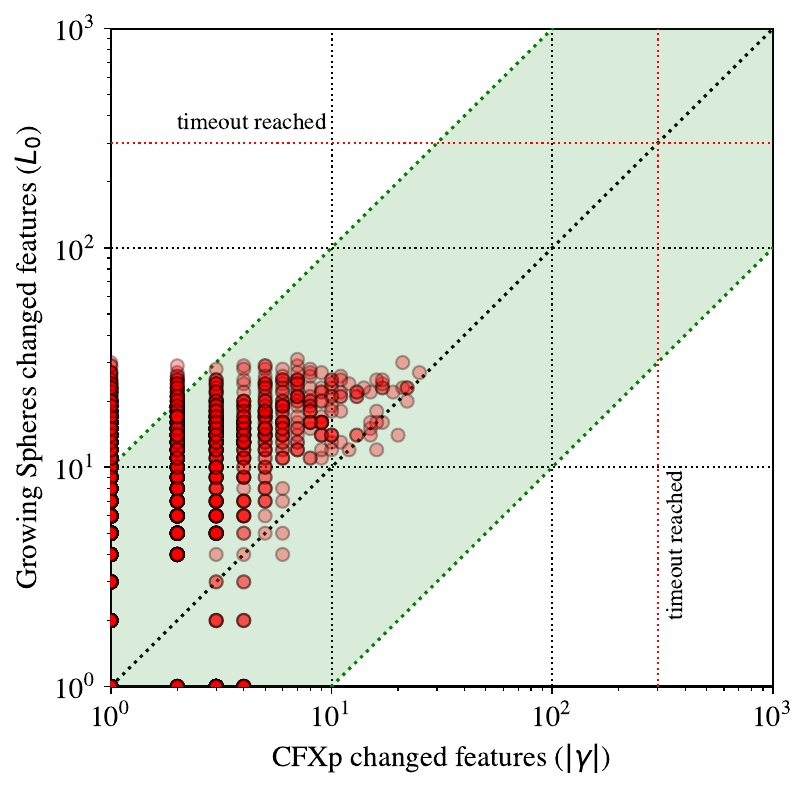}
  \caption{Growing Spheres.}
\end{subfigure}\hfill
\begin{subfigure}{.32\textwidth}
  \includegraphics[width=\linewidth]{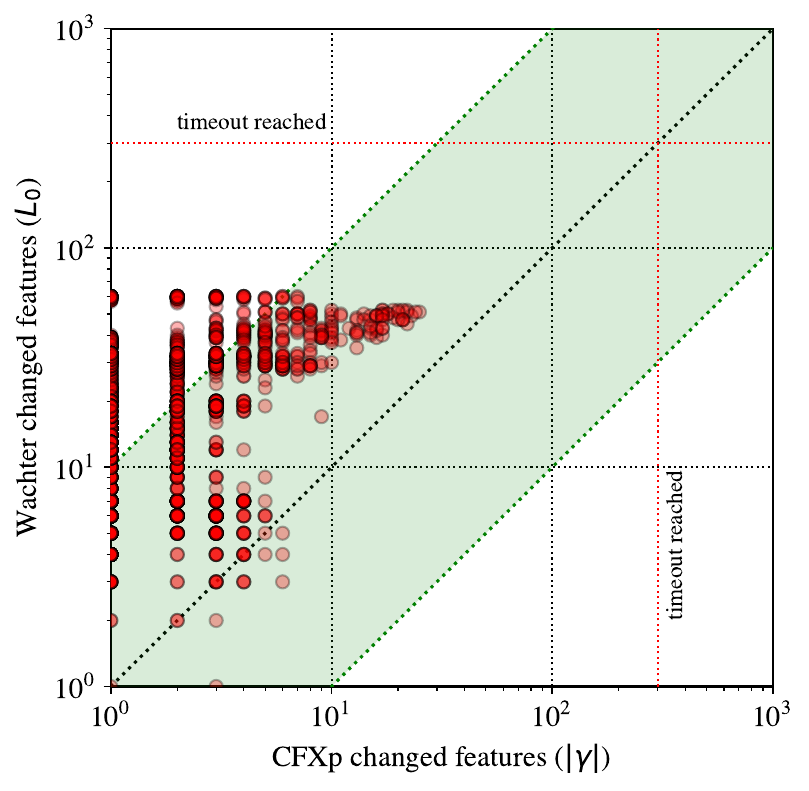}
  \caption{Wachter.}
\end{subfigure}
\caption{Comparison of explanation size for ReLU benchmarks. Only
  model-verified point outputs are plotted; the horizontal axis is
  $|\gamma|$.}
\label{fig:relu-point-sizes}
\end{figure*}

\begin{revblock}
\subsection{Additional Per-Dataset Results and Robustness of the Evaluation}
\label{sec:app-robust}

\paragraph{Per-dataset comparison with point-based explainers.}
\autoref{tab:app-t2-xgb} and \autoref{tab:app-t2-relu} detail expanded
per-dataset versions of \autoref{tab:point-res} for XGBoost and ReLU
models, respectively.
Also, \autoref{tab:app-points} reports per-dataset percentage of cases
in which the CFXp changes no more features than the counterfactual
point returned by DiCE, Growing Spheres, and Wachter (see the column
CFXp$\leq$ of \autoref{tab:point-res}).
Observe that the aggregate results against Growing Spheres and Wachter
shown earlier hold on every single dataset.
DiCE performs better than either Growing Spheres or Wachter as it
explicitly optimizes for sparsity, so CFXps change no more features in
at least half of the cases on 20 of 21 datasets for tree ensembles and
on 17 of 21 datasets for ReLU networks.
Importantly, we observe no trivial dependence of the results on the
number of features or classes in the benchmarks used: for ReLU
networks, the lowest rates occur both on a small multiclass dataset
``zoo'' and on a large binary dataset ``promoters''.

\paragraph{On CFXp regions.}
\autoref{tab:app-free} details the percentage of free features in the
CFXps per dataset.
For XGBoost, it also reports the number of features used by the model
itself and the percentage of \emph{these} features that are free in
the CFXps.
Note that for ReLU networks, 35.1\% of the original witnesses computed
by the MILP solver lie near the decision boundary (they have a score
margin approximately equal to the oracle threshold $10^{-4}$).

\begin{table}[!ht]
\centering
\small
\color{revcolor}
\captionsetup{labelfont={color=revcolor}}
\caption{\color{revcolor}Per-dataset version of \autoref{tab:point-res} for XGBoost
models. The columns are: numbers of explained instances ($n$) and
model-verified points (Ver.), mean $L_0$ distance of the computed
points, and mean $|\gamma|$ of the paired CFXps.}

\label{tab:app-t2-xgb}
\begin{adjustbox}{max width=\textwidth,center}
\begin{tabular}{lrrrrrrrrrr}
\toprule
& & \multicolumn{3}{c}{DiCE} & \multicolumn{3}{c}{Growing Spheres} & \multicolumn{3}{c}{Wachter} \\
\cmidrule(lr){3-5}\cmidrule(lr){6-8}\cmidrule(lr){9-11}
Dataset & $n$ & Ver. & $L_0$ & $|\gamma|$ & Ver. & $L_0$ & $|\gamma|$ & Ver. & $L_0$ & $|\gamma|$ \\
\midrule
ann-thyroid & 200 & 200 & 1.60 & 1.29 & 200 & 1.33 & 1.29 & 200 & 6.33 & 1.29 \\
appendicitis & 106 & 106 & 2.34 & 2.06 & 106 & 2.77 & 2.06 & 100 & 6.92 & 2.07 \\
biodegradation & 200 & 200 & 3.60 & 2.67 & 200 & 4.49 & 2.67 & 200 & 22.77 & 2.67 \\
divorce & 150 & 150 & 5.64 & 3.37 & 150 & 3.75 & 3.37 & 109 & 45.56 & 3.18 \\
ecoli & 200 & 200 & 1.46 & 1.00 & 200 & 2.59 & 1.00 & 200 & 5.20 & 1.00 \\
glass2 & 162 & 162 & 1.82 & 1.30 & 162 & 2.10 & 1.30 & 162 & 7.62 & 1.30 \\
ionosphere & 200 & 200 & 2.31 & 2.76 & 200 & 4.49 & 2.76 & 200 & 30.16 & 2.76 \\
pendigits & 200 & 200 & 1.57 & 1.22 & 200 & 4.10 & 1.22 & 200 & 13.61 & 1.22 \\
promoters & 106 & 106 & 1.44 & 1.00 & 106 & 1.00 & 1.00 & 106 & 40.72 & 1.00 \\
segmentation & 200 & 200 & 1.35 & 1.00 & 200 & 2.44 & 1.00 & 200 & 15.36 & 1.00 \\
shuttle & 200 & 200 & 1.27 & 1.01 & 200 & 1.60 & 1.01 & 200 & 6.32 & 1.01 \\
sonar & 200 & 200 & 4.63 & 4.34 & 200 & 4.74 & 4.34 & 148 & 59.39 & 4.42 \\
spambase & 200 & 200 & 2.35 & 2.38 & 200 & 4.22 & 2.38 & 200 & 16.78 & 2.38 \\
texture & 200 & 200 & 1.40 & 1.05 & 200 & 4.73 & 1.05 & 200 & 36.96 & 1.05 \\
threeOf9 & 200 & 200 & 1.41 & 1.00 & 200 & 1.00 & 1.00 & 9 & 9.00 & 1.00 \\
twonorm & 200 & 200 & 2.65 & 3.46 & 200 & 7.92 & 3.46 & 200 & 19.01 & 3.46 \\
vowel & 200 & 200 & 1.07 & 1.00 & 200 & 2.83 & 1.00 & 200 & 11.36 & 1.00 \\
wdbc & 200 & 200 & 3.65 & 3.52 & 200 & 5.67 & 3.52 & 198 & 29.26 & 3.51 \\
wine-recognition & 178 & 178 & 1.69 & 1.26 & 178 & 1.83 & 1.26 & 178 & 12.42 & 1.26 \\
wpbc & 194 & 193 & 3.59 & 2.73 & 194 & 5.52 & 2.75 & 192 & 32.17 & 2.74 \\
zoo & 59 & 59 & 1.24 & 1.02 & 59 & 1.02 & 1.02 & 59 & 3.85 & 1.02 \\
\bottomrule
\end{tabular}
\end{adjustbox}
\end{table}

\begin{table}[!ht]
\centering
\small

\color{revcolor}
\captionsetup{labelfont={color=revcolor}}
\caption{\color{revcolor}Per-dataset version of \autoref{tab:point-res} for ReLU
networks. The columns are: numbers of explained instances ($n$) and
model-verified points (Ver.), mean $L_0$ distance of the computed
points, and mean $|\gamma|$ of the paired CFXps.}

\label{tab:app-t2-relu}
\begin{adjustbox}{max width=\textwidth,center}
\begin{tabular}{lrrrrrrrrrr}
\toprule
& & \multicolumn{3}{c}{DiCE} & \multicolumn{3}{c}{Growing Spheres} & \multicolumn{3}{c}{Wachter} \\
\cmidrule(lr){3-5}\cmidrule(lr){6-8}\cmidrule(lr){9-11}
Dataset & $n$ & Ver. & $L_0$ & $|\gamma|$ & Ver. & $L_0$ & $|\gamma|$ & Ver. & $L_0$ & $|\gamma|$ \\
\midrule
ann-thyroid & 200 & 200 & 1.22 & 1.98 & 200 & 1.90 & 1.98 & 200 & 6.21 & 1.98 \\
appendicitis & 106 & 106 & 2.04 & 1.88 & 106 & 5.35 & 1.88 & 58 & 6.74 & 1.53 \\
biodegradation & 200 & 200 & 1.69 & 1.15 & 199 & 10.07 & 1.15 & 200 & 22.61 & 1.15 \\
divorce & 150 & 150 & 11.74 & 10.79 & 99 & 21.66 & 8.93 & 88 & 46.67 & 11.56 \\
ecoli & 200 & 200 & 1.30 & 1.00 & 200 & 4.53 & 1.00 & 200 & 4.29 & 1.00 \\
glass2 & 162 & 162 & 1.66 & 1.18 & 162 & 5.72 & 1.18 & 161 & 7.55 & 1.18 \\
ionosphere & 200 & 200 & 2.23 & 2.50 & 200 & 13.95 & 2.50 & 200 & 29.91 & 2.50 \\
pendigits & 200 & 200 & 1.26 & 1.10 & 200 & 8.28 & 1.10 & 200 & 13.50 & 1.10 \\
promoters & 106 & 106 & 4.20 & 6.12 & 100 & 15.87 & 6.02 & 106 & 40.43 & 6.12 \\
segmentation & 200 & 200 & 1.23 & 1.15 & 199 & 8.08 & 1.15 & 200 & 15.37 & 1.15 \\
shuttle & 200 & 200 & 1.09 & 1.22 & 200 & 5.53 & 1.22 & 200 & 6.42 & 1.22 \\
sonar & 200 & 200 & 2.65 & 2.06 & 200 & 19.97 & 2.06 & 120 & 59.38 & 2.13 \\
spambase & 200 & 200 & 1.62 & 1.23 & 200 & 9.63 & 1.23 & 200 & 16.61 & 1.23 \\
texture & 200 & 200 & 1.14 & 1.09 & 200 & 13.72 & 1.09 & 200 & 36.91 & 1.09 \\
threeOf9 & 200 & 200 & 1.42 & 1.00 & 200 & 1.55 & 1.00 & 10 & 8.10 & 1.00 \\
twonorm & 200 & 200 & 2.06 & 2.13 & 200 & 14.58 & 2.13 & 199 & 18.86 & 2.13 \\
vowel & 200 & 200 & 1.02 & 1.20 & 200 & 6.50 & 1.20 & 200 & 11.21 & 1.20 \\
wdbc & 200 & 200 & 2.03 & 2.90 & 197 & 13.61 & 2.86 & 196 & 29.31 & 2.92 \\
wine-recognition & 178 & 178 & 1.76 & 1.39 & 178 & 9.06 & 1.39 & 178 & 11.98 & 1.39 \\
wpbc & 194 & 194 & 1.70 & 1.69 & 193 & 14.61 & 1.69 & 182 & 32.21 & 1.66 \\
zoo & 59 & 59 & 1.64 & 2.98 & 59 & 5.73 & 2.98 & 59 & 4.03 & 2.98 \\
\bottomrule
\end{tabular}
\end{adjustbox}
\end{table}

\begin{table}[!ht]
\centering
\small
\color{revcolor}
\captionsetup{labelfont={color=revcolor}}
\caption{\color{revcolor}Per-dataset percentage of cases in which CFXps change no more
features than DiCE (D), Growing Spheres (GS), and Wachter (W). Values
$|\varphi|$ and $|\mbb{K}|$ represent the total numbers of features and
classes in the datasets.}
\label{tab:app-points}
\begin{tabular}{lrrrrrrrr}
\toprule
& & & \multicolumn{3}{c}{XGBoost} & \multicolumn{3}{c}{ReLU} \\
\cmidrule(lr){4-6}\cmidrule(lr){7-9}
Dataset & $|\varphi|$ & $|\mbb{K}|$ & D & GS & W & D & GS & W \\
\midrule
ann-thyroid & 21 & 3 & 99.0 & 98.5 & 100 & 52.5 & 53.0 & 100 \\
appendicitis & 7 & 2 & 94.3 & 97.2 & 100 & 92.5 & 100 & 100 \\
biodegradation & 41 & 2 & 89.0 & 97.5 & 100 & 98.0 & 100 & 100 \\
divorce & 54 & 2 & 90.7 & 74.7 & 100 & 48.0 & 96.0 & 100 \\
ecoli & 7 & 5 & 100 & 100 & 100 & 100 & 100 & 100 \\
glass2 & 9 & 2 & 92.0 & 96.9 & 100 & 96.9 & 100 & 100 \\
ionosphere & 34 & 2 & 61.0 & 87.0 & 100 & 60.0 & 100 & 100 \\
pendigits & 16 & 10 & 98.5 & 100 & 100 & 97.0 & 100 & 100 \\
promoters & 58 & 2 & 100 & 100 & 100 & 18.9 & 99.0 & 100 \\
segmentation & 19 & 7 & 100 & 100 & 100 & 92.0 & 100 & 100 \\
shuttle & 9 & 7 & 100 & 100 & 100 & 79.5 & 100 & 100 \\
sonar & 60 & 2 & 64.5 & 67.0 & 100 & 86.0 & 100 & 100 \\
spambase & 57 & 2 & 71.5 & 92.5 & 100 & 95.0 & 100 & 100 \\
texture & 40 & 11 & 98.0 & 100 & 100 & 94.0 & 100 & 100 \\
threeOf9 & 9 & 2 & 100 & 100 & 100 & 100 & 100 & 100 \\
twonorm & 20 & 2 & 40.0 & 100 & 100 & 74.0 & 100 & 100 \\
vowel & 13 & 11 & 100 & 100 & 100 & 84.0 & 100 & 100 \\
wdbc & 30 & 2 & 74.0 & 95.5 & 100 & 46.5 & 100 & 100 \\
wine-recognition & 13 & 3 & 100 & 99.4 & 100 & 96.1 & 100 & 100 \\
wpbc & 33 & 2 & 80.8 & 97.4 & 100 & 77.8 & 100 & 100 \\
zoo & 16 & 7 & 100 & 100 & 100 & 25.4 & 94.9 & 84.7 \\
\midrule
total &  &  & 87.2 & 95.3 & 100 & 79.9 & 97.2 & 99.7 \\
\bottomrule
\end{tabular}
\end{table}

\begin{table}[!ht]
\centering
\small
\color{revcolor}
\captionsetup{labelfont={color=revcolor}}
\caption{\color{revcolor}Per-dataset details on the CFXp regions: mean percentage
of free features (Free) and percentage of certificates covering only
the witness (Only). $|\varphi|$ and $u$ are the numbers of features and of
features used by the XGBoost model; Free (among used) is the percentage of
free features among the used ones.}
\label{tab:app-free}
\begin{adjustbox}{max width=\textwidth,center}
\begin{tabular}{lrrrrrrrrr}
\toprule
& & & \multicolumn{3}{c}{XGBoost} & \multicolumn{2}{c}{ReLU} &
\multicolumn{2}{c}{ReLU, improved} \\
\cmidrule(lr){4-6}\cmidrule(lr){7-8}\cmidrule(lr){9-10}
Dataset & $|\varphi|$ & $u$ & Free & Free (among used) & Only & Free & Only & Free & Only \\
\midrule
ann-thyroid & 21 & 11 & 71.0 & 44.6 & 0 & 23.2 & 0.5 & 29.3 & 0.5 \\
appendicitis & 7 & 7 & 32.9 & 32.9 & 4.7 & 10.4 & 56.6 & 29.1 & 11.3 \\
biodegradation & 41 & 34 & 49.3 & 38.9 & 0 & 17.4 & 0 & 21.8 & 0 \\
divorce & 54 & 14 & 92.6 & 71.5 & 0 & 29.0 & 2.0 & 30.4 & 0.7 \\
ecoli & 7 & 6 & 51.6 & 43.6 & 0 & 26.3 & 0 & 30.9 & 0 \\
glass2 & 9 & 8 & 50.5 & 44.3 & 0 & 9.6 & 36.4 & 14.1 & 30.9 \\
ionosphere & 34 & 31 & 60.5 & 56.7 & 0 & 17.3 & 0 & 25.2 & 0 \\
pendigits & 16 & 16 & 29.8 & 29.8 & 0.5 & 18.4 & 2.0 & 22.8 & 1.5 \\
promoters & 58 & 1 & 98.3 & 0 & 0 & 26.9 & 0 & 29.6 & 0 \\
segmentation & 19 & 17 & 64.8 & 60.7 & 0 & 19.1 & 0 & 22.9 & 0 \\
shuttle & 9 & 9 & 57.4 & 57.4 & 0 & 4.3 & 71.0 & 6.3 & 66.0 \\
sonar & 60 & 50 & 68.4 & 62.0 & 0 & 4.2 & 33.0 & 8.2 & 3.5 \\
spambase & 57 & 49 & 54.4 & 46.9 & 0 & 26.1 & 0 & 37.5 & 0 \\
texture & 40 & 37 & 58.7 & 55.3 & 0 & 5.6 & 12.0 & 6.3 & 6.5 \\
threeOf9 & 9 & 3 & 88.9 & 66.7 & 0 & 81.1 & 0 & 88.9 & 0 \\
twonorm & 20 & 20 & 20.6 & 20.6 & 0.5 & 1.6 & 71.0 & 3.2 & 46.5 \\
vowel & 13 & 13 & 41.9 & 41.9 & 0 & 7.4 & 33.0 & 8.6 & 25.0 \\
wdbc & 30 & 27 & 63.1 & 59.0 & 0 & 6.0 & 39.0 & 9.3 & 15.0 \\
wine-recognition & 13 & 12 & 69.2 & 66.7 & 0 & 6.4 & 48.9 & 14.3 & 12.9 \\
wpbc & 33 & 32 & 55.2 & 53.8 & 0 & 4.9 & 45.4 & 8.7 & 13.9 \\
zoo & 16 & 13 & 62.6 & 54.0 & 0 & 27.8 & 0 & 31.7 & 0 \\
\midrule
total & & & 58.3 & 48.9 & 0.2 & 17.4 & 21.8 & 22.2 & 11.8 \\
\bottomrule
\end{tabular}
\end{adjustbox}
\end{table}

\end{revblock}

% \subsection{Limitations}\label{sec:app-limitations}

% The evaluation concerns tabular classification and a single explanation per
% query.  The extraction procedures are deterministic conditional on a trained
% model, query, and solver configuration, but the evaluation does not estimate
% variation across retraining runs or alternative train/test splits.  CWC is
% instantiated with Hamming-style similarity on the finite tree-induced
% partition; it does not represent every possible application-specific notion
% of similarity, plausibility, or actionability.  Finally, as with other
% feature-based explanations, the guarantees concern the classifier.  Causal
% or real-world actionability requires additional domain constraints.

\end{document}